\documentclass[11pt]{article}
\usepackage[final]{ACL2023}
\usepackage{times}
\usepackage{latexsym}
\usepackage[T1]{fontenc}
\usepackage[utf8]{inputenc}
\usepackage{csquotes}
\usepackage{microtype}
\usepackage{inconsolata}
\usepackage[edges]{forest}
\usetikzlibrary{arrows.meta,shapes,positioning,shadows,trees}
\usepackage{hyperref}
\usepackage{cleveref}
\usepackage{booktabs}

\def\footurl#1{\footnote{\url{#1}}}

\title{HausaNLP: Current Status, Challenges and Future Directions for Hausa Natural Language Processing}

\author{
    Shamsuddeen Hassan Muhammad$^{1,5,6}$, Ibrahim Said Ahmad$^{2,5,6}$, Idris Abdulmumin$^{3,5}$,\\ 
    \bf Falalu Ibrahim Lawan$^{4,5}$, Sukairaj Hafiz Imam$^{5,6}$, Yusuf Aliyu$^{7}$,
    \bf Sani Abdullahi Sani$^{8}$, \\ \bf Ali Usman Umar$^{9}$, Tajuddeen Gwadabe$^5$, Kenneth Church$^{2}$, Vukosi Marivate$^{3}$\\
    \footnotesize $^1$Imperial College London, $^2$Northeastern University,
    $^3$Data Science for Social Impact, University of Pretoria, \\
    \footnotesize  $^4$Kaduna State University, $^5$HausaNLP, $^6$Bayero University, Kano, $^7$Universiti Teknologi PETRONAS,\\
    \footnotesize $^8$University of the Witwatersrand, Johannesburg, $^9$Federal University of Lafia\\ 
    \footnotesize \texttt{\textbf{correspondence}: s.muhammad@imperial.ac.uk}
}

\date{}

\begin{document}

    \maketitle

\begin{abstract}


Hausa Natural Language Processing (NLP) has gained increasing attention in recent years, yet remains understudied as a low-resource language despite having over 120 million first-language (L1) and 80 million second-language (L2) speakers worldwide. While significant advances have been made in high-resource languages, Hausa NLP faces persistent challenges including limited open-source datasets and inadequate model representation. This paper presents an overview of the current state of Hausa NLP, systematically examining existing resources, research contributions, and  gaps across fundamental NLP tasks: text classification, machine translation, named entity recognition, speech recognition, and question answering. We introduce \textsc{HausaNLP}\footnote{\url{https://catalog.hausanlp.org}}, a curated catalog 
 that aggregates datasets, tools, and research works to enhance accessibility and 
drive further development. Furthermore, we discuss challenges in integrating Hausa into large language models (LLMs), addressing issues of suboptimal tokenization, and dialectal variation. Finally, we propose strategic research directions emphasizing dataset expansion, improved language modeling approaches, and strengthened community collaboration to advance Hausa NLP. Our work provides both a foundation for accelerating Hausa NLP progress and valuable insights for broader multilingual NLP research.

\end{abstract}

\section{Introduction}

\begin{displayquote}
    \textit{The limits of my language mean the limits of my world.} -- \cite{wittgenstein1994tractatus}
\end{displayquote}

Natural Language Processing (NLP) has made significant progress and revolutionized the way language technology is used in our daily lives. From voice assistants and chatbots to machine translations, text classification, information extraction, and question-answering, NLP enables us to interact with machines in a more natural way \cite{cambria2014jumping}. 
One of the recent advances in NLP is emergence of large language models (LLMs) such as ChatGPT, which demonstrated impressive performance in various NLP tasks, such as dialogue generation and arithmetic reasoning \cite{qin2023chatgpt}. However, much of this progress has been concentrated on a limited set of high-resource languages (e.g., English and Chinese), where large-scale pre-training corpora are readily available (van Esch et al., 2022). As a result, many languages remain underrepresented in NLP research, including Hausa.

\begin{figure}
    \centering
    \resizebox{\linewidth}{!}{
    \includegraphics[scale=0.36]{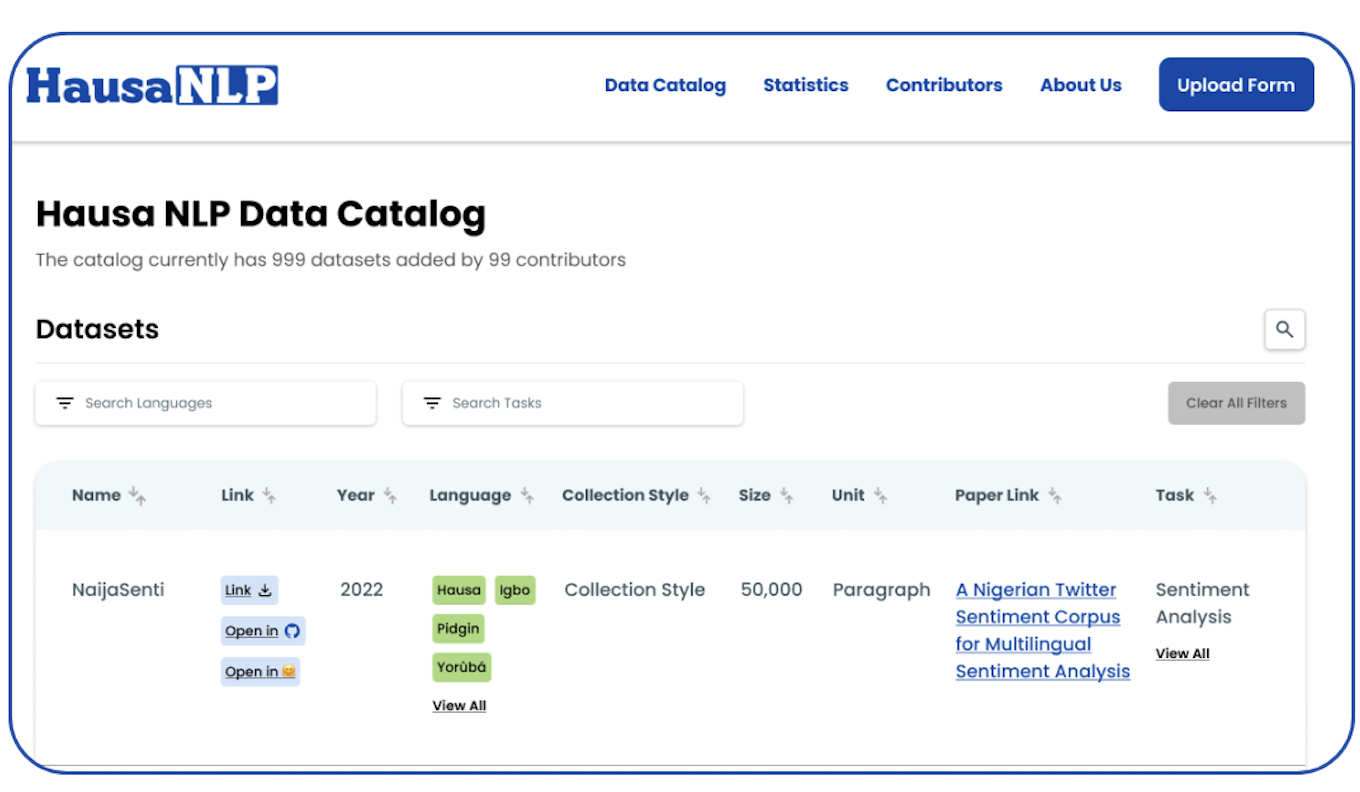}}
    \caption{\textbf{HausaNLP Catalogue:} A repository of datasets, tools, and research papers on Hausa NLP, developed to improve access to and discovery of Hausa language resources}    
    \label{fig:enter-label}
\end{figure}

Hausa is a major Chadic language with rich linguistic and cultural significance within the Afroasiatic family. Originally written in Arabic script (Ajami) during the pre-colonial era, the language has been romanized and now uses the Latin script as its primary writing system. Yet, Arabic influence remains evident in Hausa through loanwords from Arabic \cite{el1987provenance,newman_2022}. Most Hausa speakers are found in northern Nigeria and southern Niger. However, its influence has expanded through trade and migration, reaching countries such as Cameroon, Ghana, Benin, Togo, Chad, and Sudan \cite{inuwadutse2021large}. Hausa has a global presence and is broadcast by several international media outlets such as BBC, Deutsche Welle, Voice of America, Voice of Russia, China Radio International, and Radio France Internationale in Hausa ---\textit{the most predominant language broadcast internationally in West Africa}. 

Despite its importance, diversity, and cultural heritage, 
Hausa has received  relatively little attention in NLP research 
\cite{Zakari2021ASL,muhammad2025brighter, parida-etal-2023-havqa}. This slows progress in language technology research and development in Hausa and further widens the gap. Recent work on HausaNLP is mostly community-driven efforts such as machine translation \cite{adelani-etal-2022-thousand,abdulmumin-etal-2022-hausa}, sentiment analysis \cite{muhammad-etal-2022-naijasenti,muhammad2023afrisenti}, emotion detection \cite{muhammad2025brighter}, hate speech detection \cite{Muhammad2025AfriHateAM}, and named entity recognition \cite{adelani-etal-2022-masakhaner}. However, numerous NLP tasks for Hausa remain understudied, primarily due to the lack of available corpora.

Open-source corpora are key drivers of advancements in NLP. However, Hausa, a well-documented language, lacks open-source corpora that can be used for many NLP tasks. Further, the few available Hausa corpora are dispersed and difficult to access. Therefore, creating and aggregating open-source corpora for Hausa is crucial for the progress of HausaNLP. To address these challenges, this paper makes the following contributions:  


     




\begin{itemize}
    \item \textbf{HausaNLP Catalogue}: We introduce \href{https://catalog.hausanlp.org}{HausaNLP Catalogue}, a centralized repository of datasets, tools, and research papers designed to improve accessibility and accelerate progress in Hausa NLP research.
    
    \item \textbf{Comprehensive Review}: We present a review of Hausa NLP research, analyzing current progress and identifying key challenges in the field.
    
    \item \textbf{Future Directions}: We explore promising research opportunities and outline recommendations to advance Hausa NLP technologies.
\end{itemize}

We release the HausaNLP Catalogue as an open, community-driven platform to centralize and accelerate Hausa NLP research. The catalogue serves as a living resource for discovering and sharing datasets, tools, and papers, with ongoing contributions from researchers and practitioners worldwide.


\section{Hausa Language}

Hausa is the language of the Hausa people (\textit{Hausawa}), primarily spoken in West Africa's sub-Saharan region, with the largest populations in northern Nigeria and southern Niger. Significant Hausa-speaking communities exist across Northern Ghana, Togo, Cameroon, and parts of Sudan, Chad, Mali, Ivory Coast, Libya, Saudi Arabia, and the Central African Republic \citep{bello2015}. With approximately 120 million first-language (L1) and 80 million second-language (L2) speakers, Hausa ranks among Africa's most widely spoken languages, second only to Swahili in total speaker count \citep{hegazylexical}.

While some argue that Hausa may surpass Swahili in total speakers \citep{newman_2022}, Swahili maintains broader institutional recognition as an official language in four East African nations: Tanzania, Kenya, Uganda, and Rwanda. In contrast, Hausa had limited official recognition until recently, when Niger declared it an official language \citep{el1987provenance}.


Linguistically, Hausa belongs to the Chadic branch of the Afroasiatic language family and is spoken by over 200 million people either as a first language or as a second language, making it a prominent lingua franca in the region \cite{yakasai2025tauraruwaarshen}. Hausa has several dialect variations, which are broadly categorized into two major groups: western and eastern dialects. Furthermore, Hausa has regional variations influenced by contact with non-Hausa languages, leading to phonological, morphological, syntactic, and lexical differences \cite{bello2015}.

Phonologically, Hausa is a tonal language with three pitch contrasts that distinguish word meanings and grammatical categories. It has 48 phonemes and 36 standard alphabets \cite{caron2012hausa}. Morphologically, Hausa uses root-and-pattern templates and affixation to support complex morphological processes
including inflection, derivation, modification, reduplication, clipping, blending, and compounding. It also has  numerous loanwords from contact language such as Arabic \cite{ahmed1970}. Syntactically, Hausa follows a subject-verb-object (SVO) word order and uses diverse typological constructions. The language has developed two writing systems: Ajami (Arabic-based script) and Boko (Latin-based script), both actively used in print, broadcasting, and digital media.

Despite its linguistic richness, Hausa remains a low-resource language in NLP due to limited annotated corpora and tools, hindering the development of language technologies.





\begin{table*}[t]
    \centering
    \caption{Publicly available Hausa datasets}
    \renewcommand{\arraystretch}{1.5}
    \resizebox{\textwidth}{!}{
    \begin{tabular}{cp{2cm}p{2cm}p{2cm}rp{6cm}}
        \hline
        \textbf{SN} & \textbf{Source} & \textbf{Domain} & \textbf{Task} & \textbf{Size} & \textbf{Repository}  \\
        \hline
        1 & \cite{muhammad-etal-2022-naijasenti} & Tweets & Sentiment Analysis & 30k & \url{https://github.com/hausanlp/NaijaSenti/blob/main/README.md} \\
        2 & \citet{rakhmanov-schlippe-2022-sentiment} & Teachers' evaluation & Sentiment Analysis & 40k & \url{https://github.com/MrLachin/HESAC}\\
        3 & \cite{HERDPhobia} & Tweets & Hate speech detection & 6k & \url{https://github.com/hausanlp/HERDPhobia} \\
        3 & \citet{adelani2023masakhanews} & News & Topic classification & 3k & \url{https://github.com/masakhane-io/masakhane-news}\\
        4 & \cite{inuwadutse2021large}  & Tweets/News  & Machine translation, raw texts  & & \url{https://github.com/ijdutse/hausa-corpus/tree/master} \\
        5 & \cite{dione-etal-2023-masakhapos} & News & POS tagging & 1,504 sents.  & \url{https://github.com/masakhane-io/masakhane-pos/tree/main/data/hau} \\
        6 & \cite{bichi2023graph} & News & Summarization & 113  articles & \url{https://journals.plos.org/plosone/article/file?type=supplementary&id=10.1371/journal.pone.0285376.s001} \\
        7 & \cite{ogundepo-etal-2023-cross} & Wikipedia & Question Answering & 1171 & \url{https://github.com/masakhane-io/afriqa}\\
        8 & \cite{adelani-etal-2021-masakhaner, adelani-etal-2022-masakhaner} & NER & News & 2,720 \& 8,165 & \url{https://github.com/masakhane-io/masakhane-ner/}\\
        9 & \citet{adelani-etal-2022-thousand} & Machine Translation & News & & \url{https://github.com/masakhane-io/lafand-mt/tree/main} \\
        10 & \cite{akhbardeh-etal-2021-findings} & Machine Translation & News \& Religious & Numerous & \url{https://data.statmt.org/wmt21/translation-task/} \\
        11 & \cite{goyal-etal-2022-flores} & Machine Translation & Wikimedia & $\sim$2000 & \url{https://github.com/openlanguagedata/flores}\\
        12 & \cite{vegi-etal-2022-webcrawl} & Machine Translation & Web Crawl &  & \url{https://github.com/pavanpankaj/Web-Crawl-African?tab=readme-ov-file} \\
        13 & \cite{sani2025wrote} & News & Text Classification & 5172 & \url{https://github.com/TheBangis/hausa_corpus}\\ \bottomrule
    \end{tabular} }
    \label{tab:dataset}
\end{table*}

\section{Current State of Hausa NLP}
\label{sec:state}

\begin{figure}[t!]
  \centering
  \resizebox{\linewidth}{!}{
  \vspace{-3mm} 
\begin{forest}
for tree={
    scale=0.5, 
    sibling distance=3mm, 
    level distance=5mm, 
    text width=2.3cm, 
    child anchor=west,
    parent anchor=east,
    grow'=east,
    draw,
    anchor=west,
    edge path={
        \noexpand\path[\forestoption{edge},->, >={latex}] 
            (!u.parent anchor) 
                -- +(2pt,0pt) |- 
            (.child anchor)
        \forestoption{edge label};
    }
}
[Hausa NLP, root
    [Text Summarization, onode
        [\citet{bashir2017automatic}, tnode]
        [\citet{bichi2023graph}, tnode]
    ]
    [POS Tagging, onode
        [\citet{tukur-2020}, tnode]
        [\citet{awwalu2021corpus}, tnode]
        [\citet{dione-etal-2023-masakhapos}, tnode]
    ]
    [NER, onode
        [\citet{adelani-etal-2022-masakhaner}, tnode]
        [\citet{adelani-etal-2021-masakhaner}, tnode]
        [\citet{hedderich-etal-2020-transfer}, tnode]
    ] 
    [MT, onode
        [Alignment, tnode
            [\citet{abdulmumin2023leveraging}, qnode]
            [\citet{goyal-etal-2022-flores}, qnode]
        ]
        [Multimodal, tnode
            [\citet{abdulmumin-etal-2022-hausa}, qnode]
        ]
        [Translation, tnode
            [\citet{akinfaderin2020hausamt}, qnode]
            [\citet{akhbardeh-etal-2021-findings}, qnode]
            [\citet{chen-etal-2021-university}, qnode]
            [\citet{nowakowski-dwojak-2021-adam}, qnode]
            [\citet{adelani-etal-2022-thousand}, qnode]  
            [\citet{abdulmumin-etal-2022-separating}, qnode]
        ]
    ]
    [Speech Translation, onode
        [\citet{schlippe2012hausa}, tnode]
        [\citet{abubakar2024development}, tnode]
    ]
    [Text Classification, onode
        [Sentiment Analysis, tnode
            [\citet{Abubakar2021}, qnode]
            [\citet{muhammad-etal-2022-naijasenti}, qnode]
            [\citet{rakhmanov-schlippe-2022-sentiment}, qnode]
            [\citet{sani2022sentiment}, qnode]
        ]
        [Toxicity Detection, tnode
            [\citet{zandam2023online-2023}, qnode]
        ]
        [Fake News, tnode
            [\citet{sukairaj-fakenew-2022}, qnode]
        ]
    ]
    [QA, onode
        [\citet{parida-etal-2023-havqa}, tnode]
        [\citet{ogundepo-etal-2023-cross}, tnode]
    ]
]
\end{forest}
}
\caption{Taxonomy of Hausa NLP Research Progress: Tasks and Associated Publications}
\label{fig:hausa-nlp-taxonomy}
\end{figure}
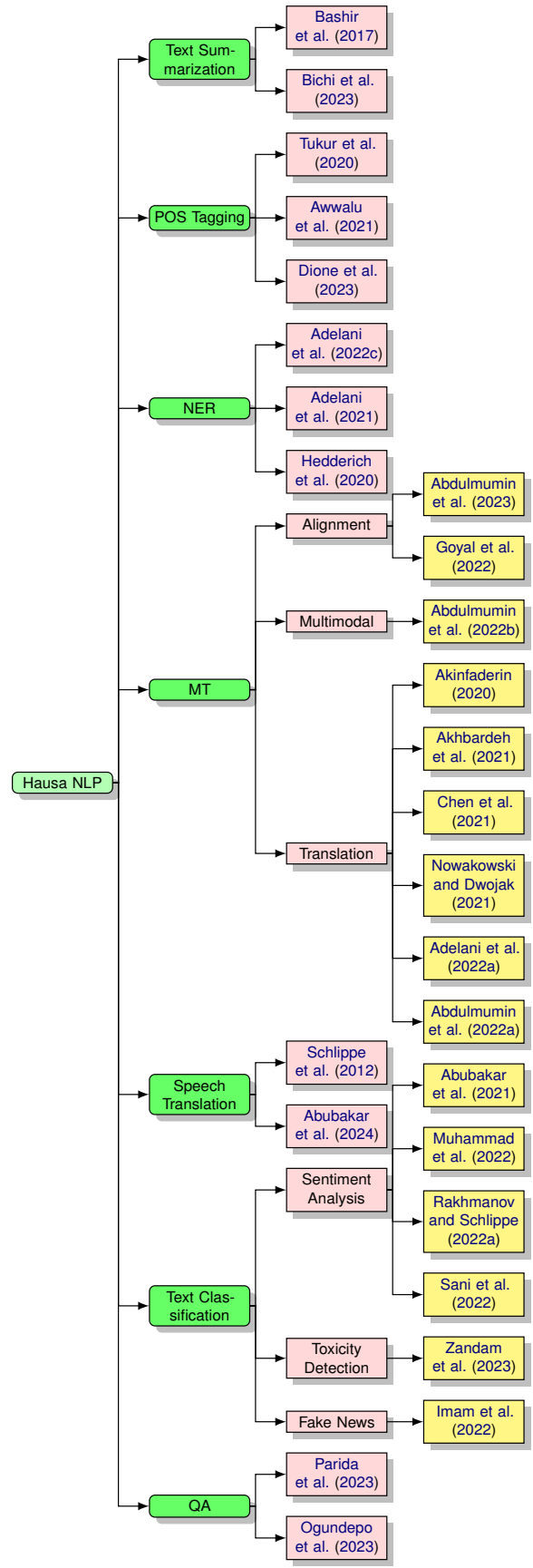

Several existing works have explored various NLP tasks in Hausa, including text classification, machine translation, named entity recognition, and automatic speech recognition, as shown in \Cref{fig:hausa-nlp-taxonomy}. This section reviews prior work on Hausa NLP, discusses available datasets, and identifies future research directions.




\tikzset{
    basic/.style  = {draw, text width=2.3cm, drop shadow, font=\sffamily, rectangle},
    root/.style   = {basic, rounded corners=2pt, thin, align=center, fill=green!30,},
    onode/.style = {basic, thin, rounded corners=2pt, align=center, fill=green!60, text width=2.3cm,},
    tnode/.style = {basic, thin, align=center, fill=pink!60, text width=2.3cm},
    qnode/.style = {basic, thin, align=center, fill=yellow!60, text width=2.3cm},
    edge from parent/.style={draw=black, edge from parent fork right}
}

\subsection{Text Classification}
Text classification is a method for automatically categorizing texts into distinct, predetermined classes. It is a supervised learning approach, as the classes must be known beforehand to train the model. Text classification can take various forms; however, in the context of Hausa texts, prior studies have primarily focused on sentiment analysis, toxicity detection, or topic classification

\paragraph{Sentiment Analysis}
Sentiment analysis is a text classification method of categorizing based on the sentiment contained in the text. The method is usually a binary classification, into positive and negative classes, or three classes, into positive, negative, and neutral classes.

Several studies have explored sentiment analysis in Hausa. \citet{Abubakar2021} introduced a sentiment analysis model for Hausa texts, leveraging a corpus of political tweets. Their approach incorporated Hausa lexical features and sentiment intensifiers, achieving an accuracy of 0.71 when employing the SVM classifier. Nevertheless, the dataset size of merely around 200 tweets in the study is grossly inadequate for training supervised learning models. 

\citet{muhammad-etal-2022-naijasenti} proposed the first large-scale sentiment dataset for the Hausa language among other Nigerian languages. The paper collected and annotated around 30,000 tweets in the Hausa language. The authors proposed novel methods for tweet collection, filtering, processing, and labeling methods. Additionally, contrary to the other study, they leverage fine-tuning LLMs, attaining a weighted F1-score of 0.81.

Further, \citet{sani2022sentiment} combined machine learning and lexicon-based approaches, achieving 86\% accuracy with TF-IDF but struggling with syntactic and semantic nuances. \citet{shehu2024unveiling} integrated CNN, RNN, and HAN with a lexicon dictionary, but the approach yielded a lower accuracy of 68.48\%, highlighting the limitations of the bag-of-words model. \citet{mohammed2024lexicon} introduced a manually annotated lexicon dataset for social media and product reviews, useful for lexicon-based models but unsuitable for data-driven approaches. To address language-specific challenges, \citet{abdullahi2024twitter} implemented a normalization process for handling Hausa abbreviations and acronyms, improving the performance of MNB and Logistic Regression. Meanwhile, \citet{ibrahim2024deep} proposed a Deep CNN model for aspect and polarity classification in Hausa movie reviews, achieving 92\% accuracy but struggling with multi-aspect classification. These studies highlight progress in Hausa sentiment analysis while emphasizing the need for better feature representation, richer datasets, and advanced techniques to handle linguistic complexities.

Future research in Hausa sentiment analysis should focus on high-quality annotated datasets to improve benchmarking \cite{liu2024application}, and domain adaptation to enhance model generalization across different contexts \cite{hays2023simplistic, singhal2023domain}, Cross-lingual sentiment classification offers potential for transferring knowledge from high-resource languages while addressing cultural nuances \cite{chan2023state, rakhmanov2022sentiment, yusuf2024sentiment}.
Further, aspect-based sentiment analysis (ABSA) is crucial for entity-level sentiment detection \cite{ibrahim2024deep, obiedat2021arabic}, while multimodal approaches integrating text, audio, and visuals remain underexplored \cite{zhu2023multimodal, gandhi2023multimodal, parida-etal-2023-havqa}. Sentiment analysis using code-mixed remains underexplored in HausaNLP \cite{shakith2024enhancing, yusuf2023fine}. Finally, explainable sentiment analysis should be explored to improve model transparency  \cite{diwali2023sentiment}. Advancing these areas will significantly strengthen Hausa NLP research and applications.

\paragraph{Emotion analysis in text}

Unlike sentiment analysis, which aims to interpret text and assign polarities (positive, negative, or neutral), emotion analysis focuses on extracting and analyzing fine-grained emotions, known as affects (e.g., happiness, sadness, fear, anger, surprise, and disgust). \citet{muhammad2025brighterbridginggaphumanannotated} is the first work on emotion detection in Hausa. The authors developed a text-based emotion dataset in 29 languages, including Hausa. The dataset is annotated into six emotion classes (anger, fear, joy, sadness, surprise, and disgust) and further categorized into intensity levels: 0 (indicating no emotion), 1 (low emotion), 2 (medium emotion), and 3 (high emotion). This dataset was used in the SemEval shared task \cite{muhammad2025brighterbridginggaphumanannotated}.



\paragraph{Toxicity detection}
Toxicity detection is a text classification task of detecting toxicity in text. The toxicity could be in the form of hate speech, harassment, and threats. The only work on toxicity detection in Hausa texts is by \cite{zandam2023online-2023}. In the work, the authors developed an online threat detection dataset using both Facebook and Twitter posts. The developed dataset is quite limited with around 801 instances. The Hausa threat detection models are based on machine learning algorithms, achieving the best performance of 0.85 with a random forest algorithm.   


\paragraph{Fake news detection}
The advancement of the internet and social media has accelerated news dissemination, offering both benefits and drawbacks. While crucial information reaches the public swiftly, the downside includes the widespread circulation of fake news. It is increasingly become difficult to distinguish actual news and fake news in the cyberspace. As a result, fake news detection has become an important area of research. 

The work of \citet{sukairaj-fakenew-2022} focused on the creation of fake news detection corpus for Hausa news articles. They developed a corpus of 2600 news articles comprising of real and fake news selected from key topics like: Business, health, entertainment, sports, politics and religion.


\paragraph{Topic Classification}
News topic classification is a text classification task in NLP that involves categorizing news articles into different categories like sports, business, entertainment, and politics. For Hausa news articles, \citet{adelani2023masakhanews} focused on topic classification for African langauges' news articles including Hausa articles. They used both classical machine learning algorithms, and pre-trained LLMs. The best performing model is AfroXLMR-large attaining a weighted F1-score of 0.92.






\subsection{Machine Translation}

\subsubsection{Text Translation}

\citet{adelani-etal-2022-thousand} leveraged pre-trained models for African news translation, focusing on 16 underrepresented African languages including the Hausa language. For the Hausa language, The Hausa Khamenei \footnote{\url{https://www.statmt.org/wmt21/translation-task.html}} corpus contained 5,898 sentences, was used. The study demonstrated the effectiveness of fine-tuning pre-trained models on a few thousand high-quality bitext for adding new languages like Hausa to the models.

\citet{nowakowski-dwojak-2021-adam} and \citet{chen-etal-2021-university} participated in the WMT 2021 News Translation Task \cite{akhbardeh-etal-2021-findings}. This involves building a machine translation system for English and Hausa language pairs. The \citet{nowakowski-dwojak-2021-adam} focused on thorough data cleaning, transfer learning, iterative training, and back-translation. The work experimented with NMT and PB-SMT, using the base Transformer architecture for the NMT models. On the other hand, \cite{chen-etal-2021-university} used an iterative back-translation approach on top of pre-trained English-German models and investigated vocabulary embedding mapping.

\citet{akinfaderin2020hausamt} explored English-Hausa machine translation by training LSTM and transformer-based model using the JW300 \cite{agic-vulic-2019-jw300} corpus. \citet{abdulmumin-etal-2022-separating} participated in WMT 2022 Large-Scale Machine Translation Evaluation for the African Languages Shared Task \cite{adelani-etal-2022-findings}. The work made an attempt to improve Hausa-English (along with other language pairs) machine translation using data filtering techniques. The idea relies on filtering out the noisy or invalid parts of a large corpus, keeping only a high-quality subset thereof. The results show that the performance of the models improved with increased data filtering, indicating the removal of noisy sentences enhanced translation quality.

\subsubsection{Multi-Modal Machine Translation}

Multimodal machine translation (MMT) focuses on translating languages using multiple modalities of information, not just text. This typically involves combining text with other data sources, such as images, speech, and video. MMT aims to enhance translation quality by incorporating information from other modalities. The goal is to leverage these additional modalities to improve the overall translation process.

\citet{abdulmumin-etal-2022-hausa} presents the \textbf{\textit{Hausa Visual Genome (HaVG)}}, a multi-modal dataset that contains the description of an image or a section within the image in Hausa and its equivalent in English. HaVG was formed by translating the English description of the images in the Hindi Visual Genome (HVG) into Hausa automatically. Afterward, the synthetic Hausa data was carefully post-edited considering the respective images. The dataset comprises 32,923 images and their descriptions.

\subsubsection{Sentence Alignment}

Automatic sentence alignment is the process of identifying which sentences in a source text correspond to which sentences in a target text. This task is crucial for creating parallel corpora, where each sentence in one language is aligned with its equivalent translation in another language. Various approaches, including length-based, lexicon-based, and translation-based methods, are employed for sentence alignment. Evaluating alignment quality involves assessing accuracy and effectiveness, considering factors like language pairs and genre. 

\citet{abdulmumin2023leveraging} addresses the challenge of limited qualitative datasets for English-Hausa machine translation by automatic sentence alignment. The work presented a qualitative parallel sentence aligner that leverages the closed-access Cohere multilingual embedding \footnote{\url{https://docs.cohere.com/docs/multilingual-language-models}}. For evaluation, the work used the MAFAND-MT \cite{adelani-etal-2022-thousand}, FLORES \cite{goyal-etal-2022-flores}, a new corpus of 1000 Hausa and English news articles each. The proposed method showed promising results. 




\subsection{POS}

Part-of-speech tagging (POS) is one of the first steps in NLP that involves the tagging (or labeling) of each word in a sentence with the correct part of speech to indicate their grammatical behaviours for computational tasks \cite{pos-martinez-2012}. POS tagging is very crucial in many NLP tasks like sentiment analysis and information extraction. 

While considerable amount of work has been done on POS tagging, only  a couple of studies are on Hausa POS tagging. \citet{tukur-2020} proposed a technique for POS tagging of Hausa sentences using the Hidden Markov Model. They evaluated the model using a manually collected and annotated Hausa corpus sourced from from radio stations. While the study is worthwhile, both the dataset and model are not publicly available. 

\citet{awwalu2021corpus} presents a study on Corpus Based Transformation-Based Learning for Hausa language POS tagging. The research involves corpus development for Hausa language POS tagset. Various models and techniques such as Transformation-Based Learning (TBL), Hidden Markov Model (HMM), and N-Gram models are employed for POS tagging. The main findings indicate that the TBL tagger outperforms HMM and N-Gram taggers in terms of accuracy levels, showcasing the effectiveness of hybrid generative and discriminative taggers.

\citet{dione-etal-2023-masakhapos} created MasakhaPOS, a large POS dataset for 20 diverse African languages. They address the challenges of using universal dependencies (UD) guidelines for these languages, and compare different POS taggers based Conditional Random Field (CRF) and several multilingual Pretrained Language Models (PLMs). For the Hausa part of the project, the data was sourced from \textit{Kano Focus} and \textit{Freedom Radio} to a total of 1504 sentences (train: 753, test:150, and dev: 601).

\subsection{Text Summarization} 

Text summarization is the process of automatically generating a concise and coherent summary of a longer text while retaining its key information and main points \cite{summarization-2021-wafaa}. 



Text summarization plays a crucial role in various applications such as information retrieval, document summarization, news aggregation, and content recommendation systems, helping users quickly grasp the main points of lengthy documents or articles.

\cite{bashir2017automatic} perhaps conducted one the the earliest works on text summarization for Hausa langauge. The work focused on text summarization based on feature extraction using Naive Bayes model. However, the validity of the work is limited by the small data size of 10 documents from news articles, with each document containing over 600 words. The work of \cite{bichi2023graph} focus on graph-based extractive text summarization method for Hausa text. The study focus on graph-based extractive single-document summarization method for Hausa text by modifying the  PageRank algorithm using the normalized common bigrams count between adjacent sentences as the initial vertex score. They evaluated the proposed approach using a manually annotated dataset that comprises of 113 Hausa news articles on various genres. Each news article had two manually generated gold standard summaries, with the length of summaries being 20\% of the original article length. 

\subsection{Question and Answering}
Question and Answering (QA) is a branch of natural language processing (NLP) that deals with building systems that can automatically answer questions posed by humans in natural language. QA systems can be useful for various applications, such as virtual assistants, customer support, search engines, and education \cite{QA-rogers-2023}.

\citet{parida-etal-2023-havqa} developed a Hausa Visual Question Answering (VQA) dataset called \textbf{\textit{HaVQA}}. The dataset is a multi-modal dataset for visual question-answering (VQA) tasks in the Hausa language. The dataset was created by manually translating 6,022 English question-answer pairs, which are associated with 1,555 unique images from the Visual Genome dataset. The paper employed state-of-the-art language and vision models for Visual Question Answering and achieved the best performance with the Data-Efficient Image Transformers model proposed by Facebook with a WuPalmer score of 30.85.

\cite{ogundepo-etal-2023-cross} developed \textbf{\textit{AfriQA}}, a dataset for cross-lingual open-retrieval question answering for 10 African languages, including the Hausa language. The dataset was developed from Wikipedia articles and manually elicited questions. For Hausa language, the final corpus consist of 1171 instances split into 435 training, 436 development and 300 test sets. The findings of the experiments proves how challenging multilingual retrieval is even for state-of-the-art QA models.

\subsection{Named Entity Recognition}

Named entity recognition (NER) is a technique of NLP that identifies and classifies named entities in a text, such as person names, organizations, locations, and dates. NER can be useful for various tasks, such as information extraction, search engines, chatbots, and machine translation. There are different methods and tools for NER, such as dictionary-based, rule-based, machine learning-based, and hybrid systems \cite{NER-2022}.


\citet{adelani-etal-2021-masakhaner} and \citet{adelani-etal-2022-masakhaner} created the largest NER corpus for African languages titled \textbf{\textit{MasakhaNER 1.0}} and \textbf{\textit{MasakhaNER 2.0}}. MasakhaNER 1.0 covers 10 African languages, while MasakhaNER 2.0 expanded the corpus to include 10 South African languages, making a total of 20 languages. MasakhaNER 1.0 consists of 2,720 sources from VOA news while  MasakhaNER 2.0 consists of 8,165 sourced from Kano Focus and Freedom Radio news channels. Both studies explored various experiments using pretrained language models and other techniques like transfer learning and zero-shot learning.

The work of \citet{hedderich-etal-2020-transfer} 
investigates transfer learning and distant supervision with multilingual transformer models on NER and topic classification in Hausa, isiXhosa and Yoruba languages. The study show that transfer learning from a high-resource language and distant supervision are effective techniques for improving performance in low-resource settings for African languages.  

\subsection{Automatic Speech Recognition (ASR)}
Automatic speech recognition (ASR) is a technology that allows computers to convert spoken language into text. ASR can be used for various purposes, such as voice control, transcription, translation, and accessibility \cite{yu2016automatic}. 

\citet{schlippe2012hausa} focused on developing a Hausa Large Vocabulary Continuous Speech Recognition (LVCSR) system by collecting a corpus of Hausa speech data from native speakers in Cameroon and text data from prominent Hausa websites. The data collected for the study included approximately 8 hours and 44 minutes of speech data from 102 native speakers of Hausa in Cameroon. Additionally, the text corpus consists of roughly 8 million words. The study found that modeling tones and vowel lengths significantly improved recognition performance, leading to a reduction in word error rates. 

\cite{abubakar2024development} focuses on developing a diacritic-aware automatic speech recognition model for the Hausa language. The model uses a large corpus of speech data from the Mozilla Common Voice dataset, which includes a variety of diacritical words and sentences. The Whisper-large model outperforms existing models, achieving a word error rate of 4.23\% and a diacritic coverage of 92\%. It also has a precision of 98.87\%, with a 2.1\% diacritic error rate, demonstrating its effectiveness in accurately transcribing Hausa speech. However, Due to the absence of prior ASR systems specifically focused on diacritization in the Hausa language, the authors were unable to make direct comparisons with their results. This lack of benchmarks may limit the ability to fully assess the effectiveness of their proposed model against existing technologies


Future efforts should prioritize developing real-time ASR systems for continuous Hausa speech recognition, enhancing usability across everyday communication and diverse industries. Optimizing computational resources and designing efficient algorithms will enable high-performance ASR systems with reduced power requirements. 
Further, exploring ASR techniques less reliant on diacritics can broaden usability for varied contexts and users. Finally, integrating ASR with NLP and machine translation can pave the way for comprehensive tools to better serve Hausa-speaking communities.


\section{Hausa Representation in Large Language Models (LLMs)}

\paragraph{}

Large language models (LLMs) have made significant strides in supporting multilingual tasks, including those involving low-resource languages like Hausa. Multilingual models such as AfrIBERTa \citep{ogueji-etal-2021-small} mBERT \citep{devlin2019bertpretrainingdeepbidirectional}, InkubaLM \citep{tonja2024inkubalmsmalllanguagemodel} XLM-R \citep{conneau-etal-2020-unsupervised}, and BLOOM \citep{workshop2023bloom176bparameteropenaccessmultilingual} have incorporated Hausa into their training data, albeit to varying degrees. These models leverage cross-lingual transfer learning to improve performance on languages with limited resources. However, the extent of Hausa representation in these models is often constrained by the scarcity of high-quality, diverse datasets. 

\paragraph{}  
The availability and quality of training data are critical factors influencing the performance of large language models (LLMs) on Hausa language tasks. Like many low-resource languages, Hausa faces challenges such as data scarcity, representational bias, and inadequate dataset construction. Existing datasets are often limited in scale and diversity, particularly in capturing dialectal variations and informal text (e.g., social media content). \citet{sani-etal-2025-investigating} highlight these challenges, emphasizing the impact of dialectal variation and tokenization on Hausa sentiment analysis. Their findings underscore the need for more diverse and high-quality datasets to enhance model performance.   Without sufficient data, LLMs struggle to achieve robust performance in handling Hausa text, as highlighted by \citet{Zhao2024} and \citet{Acikgoz2024242}.

In addition to data scarcity, Hausa's linguistic features pose significant challenges for tokenization and language modeling. The language's rich morphology, tonal variations, and complex noun pluralization systems complicate the process of accurately representing it in LLMs. Diacritics and tonal markers, which are critical for meaning, often lead to suboptimal tokenization, resulting in poor representations of the language \citep{abubakar2024development, Jaggar2006222}. Furthermore, the dialectal diversity within Hausa adds another layer of complexity. Models trained on formal Hausa text frequently struggle to process informal or dialectal variations, as noted by \citet{sani-etal-2025-investigating}. This limits their applicability in real-world scenarios where such variations are common.

Another critical issue is bias and representation in existing LLMs. Studies comparing LLM outputs with native speaker responses have revealed discrepancies in how cultural nuances and emotional tones are captured \citep{Ahmad202498}. These biases can lead to outputs that are misaligned with the cultural and linguistic expectations of Hausa speakers, further reducing the utility of LLMs for this language. Addressing these challenges requires innovative approaches, including improved tokenization strategies, dialectal adaptation techniques, and data augmentation methods. By tackling these issues, researchers can develop more robust and inclusive models that better serve Hausa speakers and other low-resource language communities




A promising direction is the development of specialized, lightweight models tailored specifically to Hausa. These custom models could provide more accurate and efficient solutions for Hausa-specific applications \citep{Yang20241}. Additionally, federated prompt tuning offers a pathway to enhance data efficiency and facilitate mutual improvements across languages, benefiting low-resource languages like Hausa \citep{Zhao2024}. Synthetic data generation also presents a valuable opportunity to address data scarcity. By creating high-quality synthetic datasets, researchers can overcome the limitations of limited real-world data and improve the performance of the model \citep{Mahgoub202486}. Together, these approaches, ranging from architectural innovations and specialized models to federated learning and synthetic data, have the potential to significantly advance Hausa representation in LLMs, making them more robust, efficient, and culturally relevant for Hausa speakers.

\section{Conclusion}
\label{sec:future}

Advancing Hausa NLP requires a multifaceted approach that addresses both technical and community-driven challenges. Below, we outline key areas for future research and development.

\paragraph{}Future research should investigate the interplay between tokenization strategies and model initialization to optimize the learning efficiency of Hausa LLMs. Techniques inspired by the BabyLM Challenge \cite{hu-etal-2024-findings} could be adapted to Hausa, focusing on sample-efficient pretraining and developmentally plausible corpora. Such approaches could mitigate data scarcity while improving model performance, particularly in low-resource settings.

\paragraph{}
Innovative architectures that support dynamic re-tokenization based on context could significantly enhance the representation of Hausa's linguistic features. These models would adapt tokenization to better capture dialectal variations and morphological complexity, improving generalization across diverse Hausa texts. This is especially important given the language's rich morphology and tonal variations, which are often underrepresented in current models.

\paragraph{}
Building on the work of \citet{wolf-etal-2023-quantifying}, future studies could explore encoding prosodic features into embeddings to improve the contextual understanding of Hausa. Although prosody carries information beyond text, its integration could enhance model performance, particularly in low-resource settings. This approach could also facilitate better handling of tonal variations in Hausa, which are critical for accurate language representation.

\paragraph{}
Creating richer and more diverse datasets for Hausa is essential for advancing NLP applications. Future efforts should focus on curating datasets that capture both formal and informal text, as well as dialectal variations. Techniques such as data augmentation, synthetic data generation, and crowdsourcing could help address data scarcity and improve model robustness. Expanding digital resources through initiatives like web crawling and community contributions \citep{schlippe2012hausa, Ibrahim2022} will also play a crucial role.

\paragraph{}
Engaging the Hausa-speaking community in dataset creation and model evaluation is vital for ensuring that LLMs reflect the linguistic and cultural nuances of Hausa. Collaborative efforts between researchers, linguists, and native speakers could lead to more representative and inclusive models. Community-driven approaches can also help address biases and improve the cultural and emotional representation of Hausa in NLP systems \citep{Ahmad202498}.

\paragraph{}
Multilingual and cross-lingual transfer learning offers promising opportunities to leverage resources from related languages to enhance Hausa NLP. For instance, the work of \citet{tokenisers_for_african_language} demonstrates that language-specific tokenizers outperform multilingual tokenizers in tasks like sentiment and news classification for African languages. Interestingly, their findings reveal that a tokenizer trained on Swahili outperformed one trained on Hausa for Hausa-specific tasks, highlighting strong cross-linguistic connections between these languages. This suggests that shared linguistic structures and features among African languages can be harnessed to improve model performance. Future research should explore these cross-linguistic bonds further, leveraging multilingual capabilities and federated learning techniques to enhance Hausa NLP \citep{Zhao2024}.

\paragraph{}
Adapting and fine-tuning existing LLMs to better handle the unique linguistic features of Hausa is another critical area for future work \citep{Acikgoz2024242, abubakar2024development}. Additionally, addressing biases and ensuring culturally aware models will be essential for creating systems that accurately represent the emotions and nuances of the Hausa language \citep{Ahmad202498}.

\section{Acknowledgements}

We would like to express our sincere gratitude to the Hack4Impact team, Vy Nguyen, Azaan Shaikh, Benjamin Chang, Sophie Lin, Keshav Subramonian, Bhuvana Betini, Evan Lin, and Sirihaasa Nallamothu, for their contributions to the development of the HausaNLP catalogue. We appreciate their efforts in advancing resources for Hausa natural language processing.

\bibliography{anthology,custom}

\begin{thebibliography}{86}
\expandafter\ifx\csname natexlab\endcsname\relax\def\natexlab#1{#1}\fi

\bibitem[{Abdullahi et~al.(2024)Abdullahi, Ahmad, and
  Haruna}]{abdullahi2024twitter}
Habeeba~Ibraheem Abdullahi, Muhammad~Aminu Ahmad, and Khalid Haruna. 2024.
\newblock Twitter sentiment analysis for hausa abbreviations and acronyms.
\newblock \emph{Science World Journal}, 19(1):101--104.

\bibitem[{Abdulmumin et~al.(2022{\natexlab{a}})Abdulmumin, Beukman, Alabi,
  Emezue, Chimoto, Adewumi, Muhammad, Adeyemi, Yousuf, Singh, and
  Gwadabe}]{abdulmumin-etal-2022-separating}
Idris Abdulmumin, Michael Beukman, Jesujoba Alabi, Chris~Chinenye Emezue,
  Everlyn Chimoto, Tosin Adewumi, Shamsuddeen Muhammad, Mofetoluwa Adeyemi,
  Oreen Yousuf, Sahib Singh, and Tajuddeen Gwadabe. 2022{\natexlab{a}}.
\newblock \href {https://aclanthology.org/2022.wmt-1.98} {Separating grains
  from the chaff: Using data filtering to improve multilingual translation for
  low-resourced {A}frican languages}.
\newblock In \emph{Proceedings of the Seventh Conference on Machine Translation
  (WMT)}, pages 1001--1014, Abu Dhabi, United Arab Emirates (Hybrid).
  Association for Computational Linguistics.

\bibitem[{Abdulmumin et~al.(2022{\natexlab{b}})Abdulmumin, Dash, Dawud, Parida,
  Muhammad, Ahmad, Panda, Bojar, Galadanci, and
  Bello}]{abdulmumin-etal-2022-hausa}
Idris Abdulmumin, Satya~Ranjan Dash, Musa~Abdullahi Dawud, Shantipriya Parida,
  Shamsuddeen Muhammad, Ibrahim~Sa{'}id Ahmad, Subhadarshi Panda, Ond{\v{r}}ej
  Bojar, Bashir~Shehu Galadanci, and Bello~Shehu Bello. 2022{\natexlab{b}}.
\newblock \href {https://aclanthology.org/2022.lrec-1.694} {{H}ausa visual
  genome: A dataset for multi-modal {E}nglish to {H}ausa machine translation}.
\newblock In \emph{Proceedings of the Thirteenth Language Resources and
  Evaluation Conference}, pages 6471--6479, Marseille, France. European
  Language Resources Association.

\bibitem[{Abdulmumin et~al.(2023)Abdulmumin, Khalid, Muhammad, Ahmad, Aliyu,
  Sani, Abduljalil, and Hassan}]{abdulmumin2023leveraging}
Idris Abdulmumin, Auwal~Abubakar Khalid, Shamsuddeen~Hassan Muhammad,
  Ibrahim~Said Ahmad, Lukman~Jibril Aliyu, Babangida Sani, Bala~Mairiga
  Abduljalil, and Sani~Ahmad Hassan. 2023.
\newblock \href {http://arxiv.org/abs/2311.12179} {Leveraging closed-access
  multilingual embedding for automatic sentence alignment in low resource
  languages}.

\bibitem[{Abubakar et~al.(2024)Abubakar, Gupta, and
  Vekkot}]{abubakar2024development}
Abdulqahar~Mukhtar Abubakar, Deepa Gupta, and Susmitha Vekkot. 2024.
\newblock Development of a diacritic-aware large vocabulary automatic speech
  recognition for hausa language.
\newblock \emph{International Journal of Speech Technology}, 27(3):687--700.

\bibitem[{Abubakar et~al.(2021)Abubakar, Roko, Bui, and Saidu}]{Abubakar2021}
Amina~Imam Abubakar, Abubakar Roko, Aminu~Muhammad Bui, and Ibrahim Saidu.
  2021.
\newblock \href {https://doi.org/10.14569/IJACSA.2021.0120913} {An enhanced
  feature acquisition for sentiment analysis of english and hausa tweets}.
\newblock \emph{International Journal of Advanced Computer Science and
  Applications}, 12(9).

\bibitem[{Acikgoz et~al.(2024)Acikgoz, Erdogan, and Yuret}]{Acikgoz2024242}
Emre~Can Acikgoz, Mete Erdogan, and Deniz Yuret. 2024.
\newblock \href
  {https://www.scopus.com/inward/record.uri?eid=2-s2.0-85216584206&partnerID=40&md5=90ee9073dd5b9fcf7bdc8cfc5440bed3}
  {Bridging the bosphorus: Advancing turkish large language models through
  strategies for low-resource language adaptation and benchmarking}.
\newblock page 242 – 268.

\bibitem[{Adelani et~al.(2022{\natexlab{a}})Adelani, Alabi, Fan, Kreutzer,
  Shen, Reid, Ruiter, Klakow, Nabende, Chang, Gwadabe, Sackey, Dossou, Emezue,
  Leong, Beukman, Muhammad, Jarso, Yousuf, Niyongabo~Rubungo, Hacheme,
  Wairagala, Nasir, Ajibade, Ajayi, Gitau, Abbott, Ahmed, Ochieng, Aremu,
  Ogayo, Mukiibi, Ouoba~Kabore, Kalipe, Mbaye, Tapo, Memdjokam~Koagne,
  Munkoh-Buabeng, Wagner, Abdulmumin, Awokoya, Buzaaba, Sibanda, Bukula, and
  Manthalu}]{adelani-etal-2022-thousand}
David Adelani, Jesujoba Alabi, Angela Fan, Julia Kreutzer, Xiaoyu Shen, Machel
  Reid, Dana Ruiter, Dietrich Klakow, Peter Nabende, Ernie Chang, Tajuddeen
  Gwadabe, Freshia Sackey, Bonaventure F.~P. Dossou, Chris Emezue, Colin Leong,
  Michael Beukman, Shamsuddeen Muhammad, Guyo Jarso, Oreen Yousuf, Andre
  Niyongabo~Rubungo, Gilles Hacheme, Eric~Peter Wairagala, Muhammad~Umair
  Nasir, Benjamin Ajibade, Tunde Ajayi, Yvonne Gitau, Jade Abbott, Mohamed
  Ahmed, Millicent Ochieng, Anuoluwapo Aremu, Perez Ogayo, Jonathan Mukiibi,
  Fatoumata Ouoba~Kabore, Godson Kalipe, Derguene Mbaye, Allahsera~Auguste
  Tapo, Victoire Memdjokam~Koagne, Edwin Munkoh-Buabeng, Valencia Wagner, Idris
  Abdulmumin, Ayodele Awokoya, Happy Buzaaba, Blessing Sibanda, Andiswa Bukula,
  and Sam Manthalu. 2022{\natexlab{a}}.
\newblock \href {https://doi.org/10.18653/v1/2022.naacl-main.223} {A few
  thousand translations go a long way! leveraging pre-trained models for
  {A}frican news translation}.
\newblock In \emph{Proceedings of the 2022 Conference of the North American
  Chapter of the Association for Computational Linguistics: Human Language
  Technologies}, pages 3053--3070, Seattle, United States. Association for
  Computational Linguistics.

\bibitem[{Adelani et~al.(2022{\natexlab{b}})Adelani, Alam, Anastasopoulos,
  Bhagia, Costa-juss{\`a}, Dodge, Faisal, Federmann, Fedorova, Guzm{\'a}n,
  Koshelev, Maillard, Marivate, Mbuya, Mourachko, Saleem, Schwenk, and
  Wenzek}]{adelani-etal-2022-findings}
David Adelani, Md~Mahfuz~Ibn Alam, Antonios Anastasopoulos, Akshita Bhagia,
  Marta~R. Costa-juss{\`a}, Jesse Dodge, Fahim Faisal, Christian Federmann,
  Natalia Fedorova, Francisco Guzm{\'a}n, Sergey Koshelev, Jean Maillard,
  Vukosi Marivate, Jonathan Mbuya, Alexandre Mourachko, Safiyyah Saleem, Holger
  Schwenk, and Guillaume Wenzek. 2022{\natexlab{b}}.
\newblock \href {https://aclanthology.org/2022.wmt-1.72} {Findings of the
  {WMT}{'}22 shared task on large-scale machine translation evaluation for
  {A}frican languages}.
\newblock In \emph{Proceedings of the Seventh Conference on Machine Translation
  (WMT)}, pages 773--800, Abu Dhabi, United Arab Emirates (Hybrid). Association
  for Computational Linguistics.

\bibitem[{Adelani et~al.(2022{\natexlab{c}})Adelani, Neubig, Ruder, Rijhwani,
  Beukman, Palen-Michel, Lignos, Alabi, Muhammad, Nabende, Dione, Bukula,
  Mabuya, Dossou, Sibanda, Buzaaba, Mukiibi, Kalipe, Mbaye, Taylor, Kabore,
  Emezue, Aremu, Ogayo, Gitau, Munkoh-Buabeng, Memdjokam~Koagne, Tapo, Macucwa,
  Marivate, Elvis, Gwadabe, Adewumi, Ahia, Nakatumba-Nabende, Mokono, Ezeani,
  Chukwuneke, Oluwaseun~Adeyemi, Hacheme, Abdulmumin, Ogundepo, Yousuf, Moteu,
  and Klakow}]{adelani-etal-2022-masakhaner}
David Adelani, Graham Neubig, Sebastian Ruder, Shruti Rijhwani, Michael
  Beukman, Chester Palen-Michel, Constantine Lignos, Jesujoba Alabi,
  Shamsuddeen Muhammad, Peter Nabende, Cheikh M.~Bamba Dione, Andiswa Bukula,
  Rooweither Mabuya, Bonaventure F.~P. Dossou, Blessing Sibanda, Happy Buzaaba,
  Jonathan Mukiibi, Godson Kalipe, Derguene Mbaye, Amelia Taylor, Fatoumata
  Kabore, Chris~Chinenye Emezue, Anuoluwapo Aremu, Perez Ogayo, Catherine
  Gitau, Edwin Munkoh-Buabeng, Victoire Memdjokam~Koagne, Allahsera~Auguste
  Tapo, Tebogo Macucwa, Vukosi Marivate, Mboning~Tchiaze Elvis, Tajuddeen
  Gwadabe, Tosin Adewumi, Orevaoghene Ahia, Joyce Nakatumba-Nabende, Neo~Lerato
  Mokono, Ignatius Ezeani, Chiamaka Chukwuneke, Mofetoluwa Oluwaseun~Adeyemi,
  Gilles~Quentin Hacheme, Idris Abdulmumin, Odunayo Ogundepo, Oreen Yousuf,
  Tatiana Moteu, and Dietrich Klakow. 2022{\natexlab{c}}.
\newblock \href {https://doi.org/10.18653/v1/2022.emnlp-main.298}
  {{M}asakha{NER} 2.0: {A}frica-centric transfer learning for named entity
  recognition}.
\newblock In \emph{Proceedings of the 2022 Conference on Empirical Methods in
  Natural Language Processing}, pages 4488--4508, Abu Dhabi, United Arab
  Emirates. Association for Computational Linguistics.

\bibitem[{Adelani et~al.(2021)Adelani, Abbott, Neubig, D{'}souza, Kreutzer,
  Lignos, Palen-Michel, Buzaaba, Rijhwani, Ruder, Mayhew, Azime, Muhammad,
  Emezue, Nakatumba-Nabende, Ogayo, Anuoluwapo, Gitau, Mbaye, Alabi, Yimam,
  Gwadabe, Ezeani, Niyongabo, Mukiibi, Otiende, Orife, David, Ngom, Adewumi,
  Rayson, Adeyemi, Muriuki, Anebi, Chukwuneke, Odu, Wairagala, Oyerinde, Siro,
  Bateesa, Oloyede, Wambui, Akinode, Nabagereka, Katusiime, Awokoya, MBOUP,
  Gebreyohannes, Tilaye, Nwaike, Wolde, Faye, Sibanda, Ahia, Dossou, Ogueji,
  DIOP, Diallo, Akinfaderin, Marengereke, and
  Osei}]{adelani-etal-2021-masakhaner}
David~Ifeoluwa Adelani, Jade Abbott, Graham Neubig, Daniel D{'}souza, Julia
  Kreutzer, Constantine Lignos, Chester Palen-Michel, Happy Buzaaba, Shruti
  Rijhwani, Sebastian Ruder, Stephen Mayhew, Israel~Abebe Azime, Shamsuddeen~H.
  Muhammad, Chris~Chinenye Emezue, Joyce Nakatumba-Nabende, Perez Ogayo, Aremu
  Anuoluwapo, Catherine Gitau, Derguene Mbaye, Jesujoba Alabi, Seid~Muhie
  Yimam, Tajuddeen~Rabiu Gwadabe, Ignatius Ezeani, Rubungo~Andre Niyongabo,
  Jonathan Mukiibi, Verrah Otiende, Iroro Orife, Davis David, Samba Ngom, Tosin
  Adewumi, Paul Rayson, Mofetoluwa Adeyemi, Gerald Muriuki, Emmanuel Anebi,
  Chiamaka Chukwuneke, Nkiruka Odu, Eric~Peter Wairagala, Samuel Oyerinde,
  Clemencia Siro, Tobius~Saul Bateesa, Temilola Oloyede, Yvonne Wambui, Victor
  Akinode, Deborah Nabagereka, Maurice Katusiime, Ayodele Awokoya, Mouhamadane
  MBOUP, Dibora Gebreyohannes, Henok Tilaye, Kelechi Nwaike, Degaga Wolde,
  Abdoulaye Faye, Blessing Sibanda, Orevaoghene Ahia, Bonaventure F.~P. Dossou,
  Kelechi Ogueji, Thierno~Ibrahima DIOP, Abdoulaye Diallo, Adewale Akinfaderin,
  Tendai Marengereke, and Salomey Osei. 2021.
\newblock \href {https://doi.org/10.1162/tacl_a_00416} {{M}asakha{NER}: Named
  entity recognition for {A}frican languages}.
\newblock \emph{Transactions of the Association for Computational Linguistics},
  9:1116--1131.

\bibitem[{Adelani et~al.(2023)Adelani, Masiak, Azime, Alabi, Tonja, Mwase,
  Ogundepo, Dossou, Oladipo, Nixdorf, Emezue, sana~al azzawi, Sibanda, David,
  Ndolela, Mukiibi, Ajayi, Moteu, Odhiambo, Owodunni, Obiefuna, Mohamed,
  Muhammad, Ababu, Salahudeen, Yigezu, Gwadabe, Abdulmumin, Taye, Awoyomi,
  Shode, Adelani, Abdulganiyu, Omotayo, Adeeko, Afolabi, Aremu, Samuel, Siro,
  Kimotho, Ogbu, Mbonu, Chukwuneke, Fanijo, Ojo, Awosan, Kebede, Sakayo,
  Nyatsine, Sidume, Yousuf, Oduwole, Tshinu, Kimanuka, Diko, Nxakama, Nigusse,
  Johar, Mohamed, Hassan, Mehamed, Ngabire, Jules, Ssenkungu, and
  Stenetorp}]{adelani2023masakhanews}
David~Ifeoluwa Adelani, Marek Masiak, Israel~Abebe Azime, Jesujoba Alabi,
  Atnafu~Lambebo Tonja, Christine Mwase, Odunayo Ogundepo, Bonaventure F.~P.
  Dossou, Akintunde Oladipo, Doreen Nixdorf, Chris~Chinenye Emezue, sana~al
  azzawi, Blessing Sibanda, Davis David, Lolwethu Ndolela, Jonathan Mukiibi,
  Tunde Ajayi, Tatiana Moteu, Brian Odhiambo, Abraham Owodunni, Nnaemeka
  Obiefuna, Muhidin Mohamed, Shamsuddeen~Hassan Muhammad, Teshome~Mulugeta
  Ababu, Saheed~Abdullahi Salahudeen, Mesay~Gemeda Yigezu, Tajuddeen Gwadabe,
  Idris Abdulmumin, Mahlet Taye, Oluwabusayo Awoyomi, Iyanuoluwa Shode,
  Tolulope Adelani, Habiba Abdulganiyu, Abdul-Hakeem Omotayo, Adetola Adeeko,
  Abeeb Afolabi, Anuoluwapo Aremu, Olanrewaju Samuel, Clemencia Siro, Wangari
  Kimotho, Onyekachi Ogbu, Chinedu Mbonu, Chiamaka Chukwuneke, Samuel Fanijo,
  Jessica Ojo, Oyinkansola Awosan, Tadesse Kebede, Toadoum~Sari Sakayo, Pamela
  Nyatsine, Freedmore Sidume, Oreen Yousuf, Mardiyyah Oduwole, Tshinu Tshinu,
  Ussen Kimanuka, Thina Diko, Siyanda Nxakama, Sinodos Nigusse, Abdulmejid
  Johar, Shafie Mohamed, Fuad~Mire Hassan, Moges~Ahmed Mehamed, Evrard Ngabire,
  Jules Jules, Ivan Ssenkungu, and Pontus Stenetorp. 2023.
\newblock \href {http://arxiv.org/abs/2304.09972} {Masakhanews: News topic
  classification for african languages}.

\bibitem[{Agi{\'c} and Vuli{\'c}(2019)}]{agic-vulic-2019-jw300}
{\v{Z}}eljko Agi{\'c} and Ivan Vuli{\'c}. 2019.
\newblock \href {https://doi.org/10.18653/v1/P19-1310} {{JW}300: A
  wide-coverage parallel corpus for low-resource languages}.
\newblock In \emph{Proceedings of the 57th Annual Meeting of the Association
  for Computational Linguistics}, pages 3204--3210, Florence, Italy.
  Association for Computational Linguistics.

\bibitem[{Ahmad et~al.(2024)Ahmad, Dudy, Ramachandranpillai, and
  Church}]{Ahmad202498}
Ibrahim~Said Ahmad, Shiran Dudy, Resmi Ramachandranpillai, and Kenneth Church.
  2024.
\newblock \href
  {https://www.scopus.com/inward/record.uri?eid=2-s2.0-85204391034&partnerID=40&md5=0f6f3a02b415783b5f532f19c14af969}
  {Are generative language models multicultural? a study on hausa culture and
  emotions using chatgpt}.
\newblock page 98 – 106.

\bibitem[{Ahmed and B.(1970)}]{ahmed1970}
U.~Ahmed and Dauda B. 1970.
\newblock An introduction to classical hausa and major dialects.
\newblock \emph{Norther Nigeria Publishing Company}.

\bibitem[{Akhbardeh et~al.(2021)Akhbardeh, Arkhangorodsky, Biesialska, Bojar,
  Chatterjee, Chaudhary, Costa-jussa, Espa{\~n}a-Bonet, Fan, Federmann,
  Freitag, Graham, Grundkiewicz, Haddow, Harter, Heafield, Homan, Huck,
  Amponsah-Kaakyire, Kasai, Khashabi, Knight, Kocmi, Koehn, Lourie, Monz,
  Morishita, Nagata, Nagesh, Nakazawa, Negri, Pal, Tapo, Turchi, Vydrin, and
  Zampieri}]{akhbardeh-etal-2021-findings}
Farhad Akhbardeh, Arkady Arkhangorodsky, Magdalena Biesialska, Ond{\v{r}}ej
  Bojar, Rajen Chatterjee, Vishrav Chaudhary, Marta~R. Costa-jussa, Cristina
  Espa{\~n}a-Bonet, Angela Fan, Christian Federmann, Markus Freitag, Yvette
  Graham, Roman Grundkiewicz, Barry Haddow, Leonie Harter, Kenneth Heafield,
  Christopher Homan, Matthias Huck, Kwabena Amponsah-Kaakyire, Jungo Kasai,
  Daniel Khashabi, Kevin Knight, Tom Kocmi, Philipp Koehn, Nicholas Lourie,
  Christof Monz, Makoto Morishita, Masaaki Nagata, Ajay Nagesh, Toshiaki
  Nakazawa, Matteo Negri, Santanu Pal, Allahsera~Auguste Tapo, Marco Turchi,
  Valentin Vydrin, and Marcos Zampieri. 2021.
\newblock \href {https://aclanthology.org/2021.wmt-1.1} {Findings of the 2021
  conference on machine translation ({WMT}21)}.
\newblock In \emph{Proceedings of the Sixth Conference on Machine Translation},
  pages 1--88, Online. Association for Computational Linguistics.

\bibitem[{Akinfaderin(2020)}]{akinfaderin2020hausamt}
Adewale Akinfaderin. 2020.
\newblock Hausamt v1. 0: Towards english--hausa neural machine translation.
\newblock In \emph{Proceedings of the The Fourth Widening Natural Language
  Processing Workshop}, pages 144--147.

\bibitem[{Aliyu et~al.(2022)Aliyu, Wajiga, Murtala, Muhammad, Abdulmumin, and
  Ahmad}]{HERDPhobia}
Saminu~Mohammad Aliyu, Gregory~Maksha Wajiga, Muhammad Murtala,
  Shamsuddeen~Hassan Muhammad, Idris Abdulmumin, and Ibrahim~Said Ahmad. 2022.
\newblock Herdphobia: A dataset for hate speech against fulani in nigeria.
\newblock In \emph{Seventh Widening Natural Language Processing Workshop
  (WiNLP)}.

\bibitem[{Awwalu et~al.(2021)Awwalu, Abdullahi, and
  Evwiekpaefe}]{awwalu2021corpus}
Jamilu Awwalu, Saleh~Elyakub Abdullahi, and Abraham~Eseoghene Evwiekpaefe.
  2021.
\newblock A corpus based transformation-based learning for hausa text parts of
  speech tagging.
\newblock \emph{International Journal of Computing and Digital Systems},
  10:473--490.

\bibitem[{Bashir et~al.(2017)Bashir, Rozaimee, and Isa}]{bashir2017automatic}
Muazzam Bashir, Azilawati Rozaimee, and Wan Malini~Wan Isa. 2017.
\newblock Automatic hausa languagetext summarization based on feature
  extraction using na{\"\i}ve bayes model.
\newblock \emph{World Applied Science Journal}, 35(9):2074--2080.

\bibitem[{Bello(2015)}]{bello2015}
A.~Bello. 2015.
\newblock The dialects of hausa.
\newblock \emph{Ahmadu Bello University Press.}

\bibitem[{Bichi et~al.(2023)Bichi, Samsudin, Hassan, Hasan, and
  Ado~Rogo}]{bichi2023graph}
Abdulkadir~Abubakar Bichi, Ruhaidah Samsudin, Rohayanti Hassan, Layla
  Rasheed~Abdallah Hasan, and Abubakar Ado~Rogo. 2023.
\newblock Graph-based extractive text summarization method for hausa text.
\newblock \emph{Plos one}, 18(5):e0285376.

\bibitem[{Cambria and White(2014)}]{cambria2014jumping}
Erik Cambria and Bebo White. 2014.
\newblock Jumping nlp curves: A review of natural language processing research.
\newblock \emph{IEEE Computational intelligence magazine}, 9(2):48--57.

\bibitem[{Caron(2012)}]{caron2012hausa}
Bernard Caron. 2012.
\newblock Hausa grammatical sketch.

\bibitem[{Chan et~al.(2023)Chan, Bea, Leow, Phoong, and Cheng}]{chan2023state}
Jireh Yi-Le Chan, Khean~Thye Bea, Steven Mun~Hong Leow, Seuk~Wai Phoong, and
  Wai~Khuen Cheng. 2023.
\newblock State of the art: a review of sentiment analysis based on sequential
  transfer learning.
\newblock \emph{Artificial Intelligence Review}, 56(1):749--780.

\bibitem[{Chen et~al.(2021)Chen, Helcl, Germann, Burchell, Bogoychev,
  Miceli~Barone, Waldendorf, Birch, and Heafield}]{chen-etal-2021-university}
Pinzhen Chen, Jind{\v{r}}ich Helcl, Ulrich Germann, Laurie Burchell, Nikolay
  Bogoychev, Antonio~Valerio Miceli~Barone, Jonas Waldendorf, Alexandra Birch,
  and Kenneth Heafield. 2021.
\newblock \href {https://aclanthology.org/2021.wmt-1.4} {The {U}niversity of
  {E}dinburgh{'}s {E}nglish-{G}erman and {E}nglish-{H}ausa submissions to the
  {WMT}21 news translation task}.
\newblock In \emph{Proceedings of the Sixth Conference on Machine Translation},
  pages 104--109, Online. Association for Computational Linguistics.

\bibitem[{Conneau et~al.(2020)Conneau, Khandelwal, Goyal, Chaudhary, Wenzek,
  Guzm{\'a}n, Grave, Ott, Zettlemoyer, and
  Stoyanov}]{conneau-etal-2020-unsupervised}
Alexis Conneau, Kartikay Khandelwal, Naman Goyal, Vishrav Chaudhary, Guillaume
  Wenzek, Francisco Guzm{\'a}n, Edouard Grave, Myle Ott, Luke Zettlemoyer, and
  Veselin Stoyanov. 2020.
\newblock \href {https://doi.org/10.18653/v1/2020.acl-main.747} {Unsupervised
  cross-lingual representation learning at scale}.
\newblock In \emph{Proceedings of the 58th Annual Meeting of the Association
  for Computational Linguistics}, pages 8440--8451, Online. Association for
  Computational Linguistics.

\bibitem[{Devlin et~al.(2019)Devlin, Chang, Lee, and
  Toutanova}]{devlin2019bertpretrainingdeepbidirectional}
Jacob Devlin, Ming-Wei Chang, Kenton Lee, and Kristina Toutanova. 2019.
\newblock \href {http://arxiv.org/abs/1810.04805} {Bert: Pre-training of deep
  bidirectional transformers for language understanding}.

\bibitem[{Dione et~al.(2023)Dione, Adelani, Nabende, Alabi, Sindane, Buzaaba,
  Muhammad, Emezue, Ogayo, Aremu, Gitau, Mbaye, Mukiibi, Sibanda, Dossou,
  Bukula, Mabuya, Tapo, Munkoh-Buabeng, Memdjokam~Koagne, Ouoba~Kabore, Taylor,
  Kalipe, Macucwa, Marivate, Gwadabe, Elvis, Onyenwe, Atindogbe, Adelani,
  Akinade, Samuel, Nahimana, Musabeyezu, Niyomutabazi, Chimhenga, Gotosa,
  Mizha, Agbolo, Traore, Uchechukwu, Yusuf, Abdullahi, and
  Klakow}]{dione-etal-2023-masakhapos}
Cheikh M.~Bamba Dione, David~Ifeoluwa Adelani, Peter Nabende, Jesujoba Alabi,
  Thapelo Sindane, Happy Buzaaba, Shamsuddeen~Hassan Muhammad, Chris~Chinenye
  Emezue, Perez Ogayo, Anuoluwapo Aremu, Catherine Gitau, Derguene Mbaye,
  Jonathan Mukiibi, Blessing Sibanda, Bonaventure F.~P. Dossou, Andiswa Bukula,
  Rooweither Mabuya, Allahsera~Auguste Tapo, Edwin Munkoh-Buabeng, Victoire
  Memdjokam~Koagne, Fatoumata Ouoba~Kabore, Amelia Taylor, Godson Kalipe,
  Tebogo Macucwa, Vukosi Marivate, Tajuddeen Gwadabe, Mboning~Tchiaze Elvis,
  Ikechukwu Onyenwe, Gratien Atindogbe, Tolulope Adelani, Idris Akinade,
  Olanrewaju Samuel, Marien Nahimana, Th{\'e}og{\`e}ne Musabeyezu, Emile
  Niyomutabazi, Ester Chimhenga, Kudzai Gotosa, Patrick Mizha, Apelete Agbolo,
  Seydou Traore, Chinedu Uchechukwu, Aliyu Yusuf, Muhammad Abdullahi, and
  Dietrich Klakow. 2023.
\newblock \href {https://doi.org/10.18653/v1/2023.acl-long.609}
  {{M}asakha{POS}: Part-of-speech tagging for typologically diverse {A}frican
  languages}.
\newblock In \emph{Proceedings of the 61st Annual Meeting of the Association
  for Computational Linguistics (Volume 1: Long Papers)}, pages 10883--10900,
  Toronto, Canada. Association for Computational Linguistics.

\bibitem[{Diwali et~al.(2023)Diwali, Saeedi, Dashtipour, Gogate, Cambria, and
  Hussain}]{diwali2023sentiment}
Arwa Diwali, Kawther Saeedi, Kia Dashtipour, Mandar Gogate, Erik Cambria, and
  Amir Hussain. 2023.
\newblock Sentiment analysis meets explainable artificial intelligence: A
  survey on explainable sentiment analysis.
\newblock \emph{IEEE Transactions on Affective Computing}.

\bibitem[{El-Kassas et~al.(2021)El-Kassas, Salama, Rafea, and
  Mohamed}]{summarization-2021-wafaa}
Wafaa~S. El-Kassas, Cherif~R. Salama, Ahmed~A. Rafea, and Hoda~K. Mohamed.
  2021.
\newblock \href {https://doi.org/https://doi.org/10.1016/j.eswa.2020.113679}
  {Automatic text summarization: A comprehensive survey}.
\newblock \emph{Expert Systems with Applications}, 165:113679.

\bibitem[{El-Shazly(1987)}]{el1987provenance}
Mohamed Helal Ahmed~Sheref El-Shazly. 1987.
\newblock \emph{The provenance of Arabic loan-words in Hausa: a phonological
  and semantic study}.
\newblock University of London, School of Oriental and African Studies (United
  Kingdom).

\bibitem[{Erasmo~Ndomba et~al.(2025)Erasmo~Ndomba, Edmund~Mswahili, and
  Jeong}]{tokenisers_for_african_language}
Goodwill Erasmo~Ndomba, Medard Edmund~Mswahili, and Young-Seob Jeong. 2025.
\newblock \href {https://doi.org/10.1109/ACCESS.2024.3522285} {Tokenizers for
  african languages}.
\newblock \emph{IEEE Access}, 13:1046--1054.

\bibitem[{Gandhi et~al.(2023)Gandhi, Adhvaryu, Poria, Cambria, and
  Hussain}]{gandhi2023multimodal}
Ankita Gandhi, Kinjal Adhvaryu, Soujanya Poria, Erik Cambria, and Amir Hussain.
  2023.
\newblock Multimodal sentiment analysis: A systematic review of history,
  datasets, multimodal fusion methods, applications, challenges and future
  directions.
\newblock \emph{Information Fusion}, 91:424--444.

\bibitem[{Goyal et~al.(2022)Goyal, Gao, Chaudhary, Chen, Wenzek, Ju, Krishnan,
  Ranzato, Guzm{\'a}n, and Fan}]{goyal-etal-2022-flores}
Naman Goyal, Cynthia Gao, Vishrav Chaudhary, Peng-Jen Chen, Guillaume Wenzek,
  Da~Ju, Sanjana Krishnan, Marc{'}Aurelio Ranzato, Francisco Guzm{\'a}n, and
  Angela Fan. 2022.
\newblock \href {https://doi.org/10.1162/tacl_a_00474} {The {F}lores-101
  evaluation benchmark for low-resource and multilingual machine translation}.
\newblock \emph{Transactions of the Association for Computational Linguistics},
  10:522--538.

\bibitem[{Hays et~al.(2023)Hays, Schutzman, Raghavan, Walk, and
  Zimmer}]{hays2023simplistic}
Chris Hays, Zachary Schutzman, Manish Raghavan, Erin Walk, and Philipp Zimmer.
  2023.
\newblock Simplistic collection and labeling practices limit the utility of
  benchmark datasets for twitter bot detection.
\newblock In \emph{Proceedings of the ACM web conference 2023}, pages
  3660--3669.

\bibitem[{Hedderich et~al.(2020)Hedderich, Adelani, Zhu, Alabi, Markus, and
  Klakow}]{hedderich-etal-2020-transfer}
Michael~A. Hedderich, David Adelani, Dawei Zhu, Jesujoba Alabi, Udia Markus,
  and Dietrich Klakow. 2020.
\newblock \href {https://doi.org/10.18653/v1/2020.emnlp-main.204} {Transfer
  learning and distant supervision for multilingual transformer models: A study
  on {A}frican languages}.
\newblock In \emph{Proceedings of the 2020 Conference on Empirical Methods in
  Natural Language Processing (EMNLP)}, pages 2580--2591, Online. Association
  for Computational Linguistics.

\bibitem[{Hegazy et~al.()Hegazy, Nofal, and Sayed}]{hegazylexical}
Mahmoud~Fahmi Hegazy, Mohammad~Ali Nofal, and MA~Mahmoud Sayed.
\newblock A lexical semantic error analysis of arabic-speaking hausa language
  learners.

\bibitem[{Hu et~al.(2024)Hu, Mueller, Ross, Williams, Linzen, Zhuang,
  Cotterell, Choshen, Warstadt, and Wilcox}]{hu-etal-2024-findings}
Michael~Y. Hu, Aaron Mueller, Candace Ross, Adina Williams, Tal Linzen, Chengxu
  Zhuang, Ryan Cotterell, Leshem Choshen, Alex Warstadt, and Ethan~Gotlieb
  Wilcox. 2024.
\newblock \href {https://aclanthology.org/2024.conll-babylm.1/} {Findings of
  the second {B}aby{LM} challenge: Sample-efficient pretraining on
  developmentally plausible corpora}.
\newblock In \emph{The 2nd BabyLM Challenge at the 28th Conference on
  Computational Natural Language Learning}, pages 1--21, Miami, FL, USA.
  Association for Computational Linguistics.

\bibitem[{Ibrahim et~al.(2024)Ibrahim, Zandam, Adam, and
  Musa}]{ibrahim2024deep}
Umar Ibrahim, Abubakar~Yakubu Zandam, Fatima~Muhammad Adam, and Aminu Musa.
  2024.
\newblock A deep convolutional neural network-based model for aspect and
  polarity classification in hausa movie reviews.
\newblock \emph{arXiv preprint arXiv:2405.19575}.

\bibitem[{Ibrahim et~al.(2022)Ibrahim, Mahatma, and Suleiman}]{Ibrahim2022}
Umar~Adam Ibrahim, Moussa~Boukar Mahatma, and Muhammed~Aliyu Suleiman. 2022.
\newblock \href {https://doi.org/10.1109/ITED56637.2022.10051610} {Framework
  for hausa speech recognition}.

\bibitem[{Imam et~al.(2022)Imam, Musa, and Choudhary}]{sukairaj-fakenew-2022}
Sukairaj~Hafiz Imam, Abubakar~Ahmad Musa, and Ankur Choudhary. 2022.
\newblock The first corpus for detecting fake news in hausa language.
\newblock In \emph{Emerging Technologies for Computing, Communication and Smart
  Cities}, pages 563--576, Singapore. Springer Nature Singapore.

\bibitem[{Inuwa-Dutse(2023)}]{inuwadutse2021large}
Isa Inuwa-Dutse. 2023.
\newblock \href {https://openreview.net/forum?id=K6FUc5TE-nY} {The first large
  scale collection of diverse hausa language datasets}.
\newblock In \emph{4th Workshop on African Natural Language Processing}.

\bibitem[{Jaggar(2006)}]{Jaggar2006222}
P.J. Jaggar. 2006.
\newblock \href {https://doi.org/10.1016/B0-08-044854-2/02071-X}
  {\emph{Hausa}}.

\bibitem[{Li et~al.(2022)Li, Sun, Han, and Li}]{NER-2022}
Jing Li, Aixin Sun, Jianglei Han, and Chenliang Li. 2022.
\newblock \href {https://doi.org/10.1109/TKDE.2020.2981314} {A survey on deep
  learning for named entity recognition}.
\newblock \emph{IEEE Transactions on Knowledge and Data Engineering},
  34(1):50--70.

\bibitem[{Liu et~al.(2024)Liu, Li, Zhu, Hong, Zhao, Dai, Wei, Huang, and
  Su}]{liu2024application}
Jiabei Liu, Keqin Li, Armando Zhu, Bo~Hong, Peng Zhao, Shuying Dai, Changsong
  Wei, Wenqian Huang, and Honghua Su. 2024.
\newblock Application of deep learning-based natural language processing in
  multilingual sentiment analysis.
\newblock \emph{Mediterranean Journal of Basic and Applied Sciences (MJBAS)},
  8(2):243--260.

\bibitem[{Mahgoub et~al.(2024)Mahgoub, Khoriba, and Elsabry}]{Mahgoub202486}
Abeer Mahgoub, Ghada Khoriba, and Elhassan~Anas Elsabry. 2024.
\newblock \href {https://doi.org/10.1016/j.procs.2024.10.181} {Mathematical
  problem solving in arabic: Assessing large language models}.
\newblock volume 244, page 86 – 95.

\bibitem[{Martinez(2012)}]{pos-martinez-2012}
Angel~R. Martinez. 2012.
\newblock \href {https://doi.org/https://doi.org/10.1002/wics.195}
  {Part-of-speech tagging}.
\newblock \emph{WIREs Computational Statistics}, 4(1):107--113.

\bibitem[{Mohammed and Prasad(2024)}]{mohammed2024lexicon}
Idi Mohammed and Rajesh Prasad. 2024.
\newblock Lexicon dataset for the hausa language.
\newblock \emph{Data in Brief}, 53:110124.

\bibitem[{Muhammad et~al.(2023)Muhammad, Abdulmumin, Ayele, Ousidhoum, Adelani,
  Yimam, Ahmad, Beloucif, Mohammad, Ruder, Hourrane, Jorge, Brazdil, Ali,
  David, Osei, Shehu-Bello, Lawan, Gwadabe, Rutunda, Belay, Messelle, Balcha,
  Chala, Gebremichael, Opoku, and Arthur}]{muhammad2023afrisenti}
Shamsuddeen Muhammad, Idris Abdulmumin, Abinew Ayele, Nedjma Ousidhoum, David
  Adelani, Seid Yimam, Ibrahim Ahmad, Meriem Beloucif, Saif Mohammad, Sebastian
  Ruder, Oumaima Hourrane, Alipio Jorge, Pavel Brazdil, Felermino Ali, Davis
  David, Salomey Osei, Bello Shehu-Bello, Falalu Lawan, Tajuddeen Gwadabe,
  Samuel Rutunda, Tadesse Belay, Wendimu Messelle, Hailu Balcha, Sisay Chala,
  Hagos Gebremichael, Bernard Opoku, and Stephen Arthur. 2023.
\newblock \href {https://doi.org/10.18653/v1/2023.emnlp-main.862}
  {{A}fri{S}enti: A {T}witter sentiment analysis benchmark for {A}frican
  languages}.
\newblock In \emph{Proceedings of the 2023 Conference on Empirical Methods in
  Natural Language Processing}, pages 13968--13981, Singapore. Association for
  Computational Linguistics.

\bibitem[{Muhammad et~al.(2025{\natexlab{a}})Muhammad, Abdulmumin, Ayele,
  Adelani, Ahmad, Aliyu, Onyango, Wanzare, Rutunda, Aliyu, Alemneh, Hourrane,
  Gebremichael, Ismail, Beloucif, Jibril, Bukula, Mabuya, Osei, Oppong, Belay,
  Guge, Asfaw, Chukwuneke, Rottger, Yimam, and
  Ousidhoum}]{Muhammad2025AfriHateAM}
Shamsuddeen~Hassan Muhammad, Idris Abdulmumin, Abinew~Ali Ayele, David~Ifeoluwa
  Adelani, Ibrahim~Said Ahmad, Saminu~Mohammad Aliyu, Nelson~Odhiambo Onyango,
  Lilian D.~A. Wanzare, Samuel Rutunda, Lukman~Jibril Aliyu, Esubalew Alemneh,
  Oumaima Hourrane, Hagos~Tesfahun Gebremichael, Elyas~Abdi Ismail, Meriem
  Beloucif, Ebrahim~Chekol Jibril, Andiswa Bukula, Rooweither Mabuya, Salomey
  Osei, Abigail Oppong, Tadesse~Destaw Belay, Tadesse~Kebede Guge,
  Tesfa~Tegegne Asfaw, Chiamaka~Ijeoma Chukwuneke, Paul Rottger, Seid~Muhie
  Yimam, and Nedjma~Djouhra Ousidhoum. 2025{\natexlab{a}}.
\newblock \href {https://api.semanticscholar.org/CorpusID:275515951} {Afrihate:
  A multilingual collection of hate speech and abusive language datasets for
  african languages}.
\newblock \emph{ArXiv}, abs/2501.08284.

\bibitem[{Muhammad et~al.(2022)Muhammad, Adelani, Ruder, Ahmad, Abdulmumin,
  Bello, Choudhury, Emezue, Abdullahi, Aremu, Jorge, and
  Brazdil}]{muhammad-etal-2022-naijasenti}
Shamsuddeen~Hassan Muhammad, David~Ifeoluwa Adelani, Sebastian Ruder,
  Ibrahim~Sa{'}id Ahmad, Idris Abdulmumin, Bello~Shehu Bello, Monojit
  Choudhury, Chris~Chinenye Emezue, Saheed~Salahudeen Abdullahi, Anuoluwapo
  Aremu, Al{\'\i}pio Jorge, and Pavel Brazdil. 2022.
\newblock \href {https://aclanthology.org/2022.lrec-1.63} {{N}aija{S}enti: A
  nigerian {T}witter sentiment corpus for multilingual sentiment analysis}.
\newblock In \emph{Proceedings of the Thirteenth Language Resources and
  Evaluation Conference}, pages 590--602, Marseille, France. European Language
  Resources Association.

\bibitem[{Muhammad et~al.(2025{\natexlab{b}})Muhammad, Ousidhoum, Abdulmumin,
  Wahle, Ruas, Beloucif, de~Kock, Surange, Teodorescu, Ahmad, Adelani, Aji,
  Ali, Alimova, Araujo, Babakov, Baes, Bucur, Bukula, Cao, Cardenas, Chevi,
  Chukwuneke, Ciobotaru, Dementieva, Gadanya, Geislinger, Gipp, Hourrane,
  Ignat, Lawan, Mabuya, Mahendra, Marivate, Piper, Panchenko, Ferreira,
  Protasov, Rutunda, Shrivastava, Udrea, Wanzare, Wu, Wunderlich, Zhafran,
  Zhang, Zhou, and Mohammad}]{muhammad2025brighterbridginggaphumanannotated}
Shamsuddeen~Hassan Muhammad, Nedjma Ousidhoum, Idris Abdulmumin, Jan~Philip
  Wahle, Terry Ruas, Meriem Beloucif, Christine de~Kock, Nirmal Surange,
  Daniela Teodorescu, Ibrahim~Said Ahmad, David~Ifeoluwa Adelani, Alham~Fikri
  Aji, Felermino D. M.~A. Ali, Ilseyar Alimova, Vladimir Araujo, Nikolay
  Babakov, Naomi Baes, Ana-Maria Bucur, Andiswa Bukula, Guanqun Cao,
  Rodrigo~Tufino Cardenas, Rendi Chevi, Chiamaka~Ijeoma Chukwuneke, Alexandra
  Ciobotaru, Daryna Dementieva, Murja~Sani Gadanya, Robert Geislinger, Bela
  Gipp, Oumaima Hourrane, Oana Ignat, Falalu~Ibrahim Lawan, Rooweither Mabuya,
  Rahmad Mahendra, Vukosi Marivate, Andrew Piper, Alexander Panchenko, Charles
  Henrique~Porto Ferreira, Vitaly Protasov, Samuel Rutunda, Manish Shrivastava,
  Aura~Cristina Udrea, Lilian Diana~Awuor Wanzare, Sophie Wu, Florian~Valentin
  Wunderlich, Hanif~Muhammad Zhafran, Tianhui Zhang, Yi~Zhou, and Saif~M.
  Mohammad. 2025{\natexlab{b}}.
\newblock \href {http://arxiv.org/abs/2502.11926} {Brighter: Bridging the gap
  in human-annotated textual emotion recognition datasets for 28 languages}.

\bibitem[{Muhammad et~al.(2025{\natexlab{c}})Muhammad, Ousidhoum, Abdulmumin,
  Wahle, Ruas, Beloucif, de~Kock, Surange, Teodorescu, Ahmad
  et~al.}]{muhammad2025brighter}
Shamsuddeen~Hassan Muhammad, Nedjma Ousidhoum, Idris Abdulmumin, Jan~Philip
  Wahle, Terry Ruas, Meriem Beloucif, Christine de~Kock, Nirmal Surange,
  Daniela Teodorescu, Ibrahim~Said Ahmad, et~al. 2025{\natexlab{c}}.
\newblock Brighter: Bridging the gap in human-annotated textual emotion
  recognition datasets for 28 languages.
\newblock \emph{arXiv preprint arXiv:2502.11926}.

\bibitem[{Newman(2022)}]{newman_2022}
Paul Newman. 2022.
\newblock \href {https://doi.org/10.1017/9781009128070.006} {\emph{Loanwords}},
  page 205–211. Cambridge University Press.

\bibitem[{Nowakowski and Dwojak(2021)}]{nowakowski-dwojak-2021-adam}
Artur Nowakowski and Tomasz Dwojak. 2021.
\newblock \href {https://aclanthology.org/2021.wmt-1.14} {{A}dam {M}ickiewicz
  {U}niversity{'}s {E}nglish-{H}ausa submissions to the {WMT} 2021 news
  translation task}.
\newblock In \emph{Proceedings of the Sixth Conference on Machine Translation},
  pages 167--171, Online. Association for Computational Linguistics.

\bibitem[{Obiedat et~al.(2021)Obiedat, Al-Darras, Alzaghoul, and
  Harfoushi}]{obiedat2021arabic}
Ruba Obiedat, Duha Al-Darras, Esra Alzaghoul, and Osama Harfoushi. 2021.
\newblock Arabic aspect-based sentiment analysis: A systematic literature
  review.
\newblock \emph{IEEE Access}, 9:152628--152645.

\bibitem[{Ogueji et~al.(2021)Ogueji, Zhu, and Lin}]{ogueji-etal-2021-small}
Kelechi Ogueji, Yuxin Zhu, and Jimmy Lin. 2021.
\newblock \href {https://doi.org/10.18653/v1/2021.mrl-1.11} {Small data? no
  problem! exploring the viability of pretrained multilingual language models
  for low-resourced languages}.
\newblock In \emph{Proceedings of the 1st Workshop on Multilingual
  Representation Learning}, pages 116--126, Punta Cana, Dominican Republic.
  Association for Computational Linguistics.

\bibitem[{Ogundepo et~al.(2023)Ogundepo, Gwadabe, Rivera, Clark, Ruder,
  Adelani, Dossou, Diop, Sikasote, Hacheme, Buzaaba, Ezeani, Mabuya, Osei,
  Emezue, Kahira, Muhammad, Oladipo, Owodunni, Tonja, Shode, Asai, Aremu,
  Awokoya, Opoku, Chukwuneke, Mwase, Siro, Arthur, Ajayi, Otiende, Rubungo,
  Sinkala, Ajisafe, Onwuegbuzia, Lawan, Ahmad, Alabi, Mbonu, Adeyemi, Phiri,
  Ahia, Iro, and Adhiambo}]{ogundepo-etal-2023-cross}
Odunayo Ogundepo, Tajuddeen Gwadabe, Clara Rivera, Jonathan Clark, Sebastian
  Ruder, David Adelani, Bonaventure Dossou, Abdou Diop, Claytone Sikasote,
  Gilles Hacheme, Happy Buzaaba, Ignatius Ezeani, Rooweither Mabuya, Salomey
  Osei, Chris Emezue, Albert Kahira, Shamsuddeen Muhammad, Akintunde Oladipo,
  Abraham Owodunni, Atnafu Tonja, Iyanuoluwa Shode, Akari Asai, Anuoluwapo
  Aremu, Ayodele Awokoya, Bernard Opoku, Chiamaka Chukwuneke, Christine Mwase,
  Clemencia Siro, Stephen Arthur, Tunde Ajayi, Verrah Otiende, Andre Rubungo,
  Boyd Sinkala, Daniel Ajisafe, Emeka Onwuegbuzia, Falalu Lawan, Ibrahim Ahmad,
  Jesujoba Alabi, Chinedu Mbonu, Mofetoluwa Adeyemi, Mofya Phiri, Orevaoghene
  Ahia, Ruqayya Iro, and Sonia Adhiambo. 2023.
\newblock \href {https://doi.org/10.18653/v1/2023.findings-emnlp.997}
  {Cross-lingual open-retrieval question answering for {A}frican languages}.
\newblock In \emph{Findings of the Association for Computational Linguistics:
  EMNLP 2023}, pages 14957--14972, Singapore. Association for Computational
  Linguistics.

\bibitem[{Parida et~al.(2023)Parida, Abdulmumin, Muhammad, Bose, Kohli, Ahmad,
  Kotwal, Deb~Sarkar, Bojar, and Kakudi}]{parida-etal-2023-havqa}
Shantipriya Parida, Idris Abdulmumin, Shamsuddeen~Hassan Muhammad, Aneesh Bose,
  Guneet~Singh Kohli, Ibrahim~Said Ahmad, Ketan Kotwal, Sayan Deb~Sarkar,
  Ond{\v{r}}ej Bojar, and Habeebah Kakudi. 2023.
\newblock \href {https://doi.org/10.18653/v1/2023.findings-acl.646} {{H}a{VQA}:
  A dataset for visual question answering and multimodal research in {H}ausa
  language}.
\newblock In \emph{Findings of the Association for Computational Linguistics:
  ACL 2023}, pages 10162--10183, Toronto, Canada. Association for Computational
  Linguistics.

\bibitem[{Qin et~al.(2023)Qin, Zhang, Zhang, Chen, Yasunaga, and
  Yang}]{qin2023chatgpt}
Chengwei Qin, Aston Zhang, Zhuosheng Zhang, Jiaao Chen, Michihiro Yasunaga, and
  Diyi Yang. 2023.
\newblock Is chatgpt a general-purpose natural language processing task solver?
\newblock \emph{arXiv preprint arXiv:2302.06476}.

\bibitem[{Rakhmanov and
  Schlippe(2022{\natexlab{a}})}]{rakhmanov-schlippe-2022-sentiment}
Ochilbek Rakhmanov and Tim Schlippe. 2022{\natexlab{a}}.
\newblock \href {https://aclanthology.org/2022.sigul-1.13} {Sentiment analysis
  for {H}ausa: Classifying students{'} comments}.
\newblock In \emph{Proceedings of the 1st Annual Meeting of the ELRA/ISCA
  Special Interest Group on Under-Resourced Languages}, pages 98--105,
  Marseille, France. European Language Resources Association.

\bibitem[{Rakhmanov and Schlippe(2022{\natexlab{b}})}]{rakhmanov2022sentiment}
Ochilbek Rakhmanov and Tim Schlippe. 2022{\natexlab{b}}.
\newblock Sentiment analysis for hausa: Classifying students’ comments.
\newblock In \emph{Proceedings of the 1st Annual Meeting of the ELRA/ISCA
  Special Interest Group on Under-Resourced Languages}, pages 98--105.

\bibitem[{Rogers et~al.(2023)Rogers, Gardner, and Augenstein}]{QA-rogers-2023}
Anna Rogers, Matt Gardner, and Isabelle Augenstein. 2023.
\newblock \href {https://doi.org/10.1145/3560260} {Qa dataset explosion: A
  taxonomy of nlp resources for question answering and reading comprehension}.
\newblock \emph{ACM Comput. Surv.}, 55(10).

\bibitem[{Sani et~al.(2025{\natexlab{a}})Sani, Soy, Imam, Mustapha, Aliyu,
  Abdulmumin, Ahmad, and Muhammad}]{sani2025wrote}
Babangida Sani, Aakansha Soy, Sukairaj~Hafiz Imam, Ahmad Mustapha,
  Lukman~Jibril Aliyu, Idris Abdulmumin, Ibrahim~Said Ahmad, and
  Shamsuddeen~Hassan Muhammad. 2025{\natexlab{a}}.
\newblock Who wrote this? identifying machine vs human-generated text in hausa.
\newblock \emph{arXiv preprint arXiv:2503.13101}.

\bibitem[{Sani et~al.(2022)Sani, Ahmad, and Abdulazeez}]{sani2022sentiment}
Muhammad Sani, Abubakar Ahmad, and Hadiza~S Abdulazeez. 2022.
\newblock Sentiment analysis of hausa language tweet using machine learning
  approach.
\newblock \emph{Journal of Research in Applied Mathematics}, 8(9):07--16.

\bibitem[{Sani et~al.(2025{\natexlab{b}})Sani, Muhammad, and
  Jarvis}]{sani-etal-2025-investigating}
Sani~Abdullahi Sani, Shamsuddeen~Hassan Muhammad, and Devon Jarvis.
  2025{\natexlab{b}}.
\newblock \href {https://aclanthology.org/2025.loreslm-1.7/} {Investigating the
  impact of language-adaptive fine-tuning on sentiment analysis in {H}ausa
  language using {A}fri{BERT}a}.
\newblock In \emph{Proceedings of the First Workshop on Language Models for
  Low-Resource Languages}, pages 101--111, Abu Dhabi, United Arab Emirates.
  Association for Computational Linguistics.

\bibitem[{Schlippe et~al.(2012)Schlippe, Djomgang, Vu, Ochs, and
  Schultz}]{schlippe2012hausa}
Tim Schlippe, Edy Guevara~Komgang Djomgang, Ngoc~Thang Vu, Sebastian Ochs, and
  Tanja Schultz. 2012.
\newblock Hausa large vocabulary continuous speech recognition.
\newblock In \emph{Spoken Language Technologies for Under-Resourced Languages}.

\bibitem[{Shakith and Arockiam(2024)}]{shakith2024enhancing}
Ayesha Shakith and L~Arockiam. 2024.
\newblock Enhancing classification accuracy on code-mixed and imbalanced data
  using an adaptive deep autoencoder and xgboost.
\newblock \emph{The Scientific Temper}, 15(03):2598--2608.

\bibitem[{Shehu et~al.(2024)Shehu, Majikumna, Suleiman, Luka, Sharif, Ramadan,
  and Kusetogullari}]{shehu2024unveiling}
Harisu~Abdullahi Shehu, Kaloma~Usman Majikumna, Aminu~Bashir Suleiman, Stephen
  Luka, Md~Haidar Sharif, Rabie~A Ramadan, and Huseyin Kusetogullari. 2024.
\newblock Unveiling sentiments: A deep dive into sentiment analysis for
  low-resource languages--a case study on hausa texts.
\newblock \emph{IEEE Access}.

\bibitem[{Singhal et~al.(2023)Singhal, Walambe, Ramanna, and
  Kotecha}]{singhal2023domain}
Peeyush Singhal, Rahee Walambe, Sheela Ramanna, and Ketan Kotecha. 2023.
\newblock Domain adaptation: challenges, methods, datasets, and applications.
\newblock \emph{IEEE access}, 11:6973--7020.

\bibitem[{Tonja et~al.(2024)Tonja, Dossou, Ojo, Rajab, Thior, Wairagala, Aremu,
  Moiloa, Abbott, Marivate, and Rosman}]{tonja2024inkubalmsmalllanguagemodel}
Atnafu~Lambebo Tonja, Bonaventure F.~P. Dossou, Jessica Ojo, Jenalea Rajab,
  Fadel Thior, Eric~Peter Wairagala, Anuoluwapo Aremu, Pelonomi Moiloa, Jade
  Abbott, Vukosi Marivate, and Benjamin Rosman. 2024.
\newblock \href {http://arxiv.org/abs/2408.17024} {Inkubalm: A small language
  model for low-resource african languages}.

\bibitem[{Tukur et~al.(2020)Tukur, Umar, and Muhammad}]{tukur-2020}
Aminu Tukur, Kabir Umar, and Anas~Sa’idu Muhammad. 2020.
\newblock Parts-of-speech tagging of hausa-based texts using hidden markov
  model.
\newblock \emph{Dutse Journal of Pure and Applied Sciences (DUJOPAS)},
  6:303--313.

\bibitem[{Vegi et~al.(2022)Vegi, J, Paul, Mishra, Banjare, K~R, and
  Viswanathan}]{vegi-etal-2022-webcrawl}
Pavanpankaj Vegi, Sivabhavani J, Biswajit Paul, Abhinav Mishra, Prashant
  Banjare, Prasanna K~R, and Chitra Viswanathan. 2022.
\newblock \href {https://aclanthology.org/2022.wmt-1.105} {{W}eb{C}rawl
  {A}frican : A multilingual parallel corpora for {A}frican languages}.
\newblock In \emph{Proceedings of the Seventh Conference on Machine Translation
  (WMT)}, pages 1076--1089, Abu Dhabi, United Arab Emirates (Hybrid).
  Association for Computational Linguistics.

\bibitem[{Wittgenstein(1994)}]{wittgenstein1994tractatus}
Ludwig Wittgenstein. 1994.
\newblock \emph{Tractatus logico-philosophicus}.
\newblock Edusp.

\bibitem[{Wolf et~al.(2023)Wolf, Pimentel, Fedorenko, Cotterell, Warstadt,
  Wilcox, and Regev}]{wolf-etal-2023-quantifying}
Lukas Wolf, Tiago Pimentel, Evelina Fedorenko, Ryan Cotterell, Alex Warstadt,
  Ethan Wilcox, and Tamar Regev. 2023.
\newblock \href {https://doi.org/10.18653/v1/2023.emnlp-main.606} {Quantifying
  the redundancy between prosody and text}.
\newblock In \emph{Proceedings of the 2023 Conference on Empirical Methods in
  Natural Language Processing}, pages 9765--9784, Singapore. Association for
  Computational Linguistics.

\bibitem[{Workshop et~al.(2023)Workshop, :, Scao, Fan, Akiki, Pavlick, Ilić,
  Hesslow, Castagné, Luccioni, Yvon, Gallé, Tow, Rush, Biderman, Webson,
  Ammanamanchi, Wang, Sagot, Muennighoff, del Moral, Ruwase, Bawden, Bekman,
  McMillan-Major, Beltagy, Nguyen, Saulnier, Tan, Suarez, Sanh, Laurençon,
  Jernite, Launay, Mitchell, Raffel, Gokaslan, Simhi, Soroa, Aji, Alfassy,
  Rogers, Nitzav, Xu, Mou, Emezue, Klamm, Leong, van Strien, Adelani, Radev,
  Ponferrada, Levkovizh, Kim, Natan, Toni, Dupont, Kruszewski, Pistilli,
  Elsahar, Benyamina, Tran, Yu, Abdulmumin, Johnson, Gonzalez-Dios, de~la Rosa,
  Chim, Dodge, Zhu, Chang, Frohberg, Tobing, Bhattacharjee, Almubarak, Chen,
  Lo, Werra, Weber, Phan, allal, Tanguy, Dey, Muñoz, Masoud, Grandury,
  Šaško, Huang, Coavoux, Singh, Jiang, Vu, Jauhar, Ghaleb, Subramani,
  Kassner, Khamis, Nguyen, Espejel, de~Gibert, Villegas, Henderson, Colombo,
  Amuok, Lhoest, Harliman, Bommasani, López, Ribeiro, Osei, Pyysalo, Nagel,
  Bose, Muhammad, Sharma, Longpre, Nikpoor, Silberberg, Pai, Zink, Torrent,
  Schick, Thrush, Danchev, Nikoulina, Laippala, Lepercq, Prabhu, Alyafeai,
  Talat, Raja, Heinzerling, Si, Taşar, Salesky, Mielke, Lee, Sharma, Santilli,
  Chaffin, Stiegler, Datta, Szczechla, Chhablani, Wang, Pandey, Strobelt,
  Fries, Rozen, Gao, Sutawika, Bari, Al-shaibani, Manica, Nayak, Teehan,
  Albanie, Shen, Ben-David, Bach, Kim, Bers, Fevry, Neeraj, Thakker, Raunak,
  Tang, Yong, Sun, Brody, Uri, Tojarieh, Roberts, Chung, Tae, Phang, Press, Li,
  Narayanan, Bourfoune, Casper, Rasley, Ryabinin, Mishra, Zhang, Shoeybi,
  Peyrounette, Patry, Tazi, Sanseviero, von Platen, Cornette, Lavallée,
  Lacroix, Rajbhandari, Gandhi, Smith, Requena, Patil, Dettmers, Baruwa, Singh,
  Cheveleva, Ligozat, Subramonian, Névéol, Lovering, Garrette, Tunuguntla,
  Reiter, Taktasheva, Voloshina, Bogdanov, Winata, Schoelkopf, Kalo, Novikova,
  Forde, Clive, Kasai, Kawamura, Hazan, Carpuat, Clinciu, Kim, Cheng, Serikov,
  Antverg, van~der Wal, Zhang, Zhang, Gehrmann, Mirkin, Pais, Shavrina,
  Scialom, Yun, Limisiewicz, Rieser, Protasov, Mikhailov, Pruksachatkun,
  Belinkov, Bamberger, Kasner, Rueda, Pestana, Feizpour, Khan, Faranak, Santos,
  Hevia, Unldreaj, Aghagol, Abdollahi, Tammour, HajiHosseini, Behroozi,
  Ajibade, Saxena, Ferrandis, McDuff, Contractor, Lansky, David, Kiela, Nguyen,
  Tan, Baylor, Ozoani, Mirza, Ononiwu, Rezanejad, Jones, Bhattacharya,
  Solaiman, Sedenko, Nejadgholi, Passmore, Seltzer, Sanz, Dutra, Samagaio,
  Elbadri, Mieskes, Gerchick, Akinlolu, McKenna, Qiu, Ghauri, Burynok, Abrar,
  Rajani, Elkott, Fahmy, Samuel, An, Kromann, Hao, Alizadeh, Shubber, Wang,
  Roy, Viguier, Le, Oyebade, Le, Yang, Nguyen, Kashyap, Palasciano, Callahan,
  Shukla, Miranda-Escalada, Singh, Beilharz, Wang, Brito, Zhou, Jain, Xu,
  Fourrier, Periñán, Molano, Yu, Manjavacas, Barth, Fuhrimann, Altay, Bayrak,
  Burns, Vrabec, Bello, Dash, Kang, Giorgi, Golde, Posada, Sivaraman,
  Bulchandani, Liu, Shinzato, de~Bykhovetz, Takeuchi, Pàmies, Castillo,
  Nezhurina, Sänger, Samwald, Cullan, Weinberg, Wolf, Mihaljcic, Liu,
  Freidank, Kang, Seelam, Dahlberg, Broad, Muellner, Fung, Haller,
  Chandrasekhar, Eisenberg, Martin, Canalli, Su, Su, Cahyawijaya, Garda,
  Deshmukh, Mishra, Kiblawi, Ott, Sang-aroonsiri, Kumar, Schweter, Bharati,
  Laud, Gigant, Kainuma, Kusa, Labrak, Bajaj, Venkatraman, Xu, Xu, Xu, Tan,
  Xie, Ye, Bras, Belkada, and
  Wolf}]{workshop2023bloom176bparameteropenaccessmultilingual}
BigScience Workshop, :, Teven~Le Scao, Angela Fan, Christopher Akiki, Ellie
  Pavlick, Suzana Ilić, Daniel Hesslow, Roman Castagné, Alexandra~Sasha
  Luccioni, François Yvon, Matthias Gallé, Jonathan Tow, Alexander~M. Rush,
  Stella Biderman, Albert Webson, Pawan~Sasanka Ammanamanchi, Thomas Wang,
  Benoît Sagot, Niklas Muennighoff, Albert~Villanova del Moral, Olatunji
  Ruwase, Rachel Bawden, Stas Bekman, Angelina McMillan-Major, Iz~Beltagy, Huu
  Nguyen, Lucile Saulnier, Samson Tan, Pedro~Ortiz Suarez, Victor Sanh, Hugo
  Laurençon, Yacine Jernite, Julien Launay, Margaret Mitchell, Colin Raffel,
  Aaron Gokaslan, Adi Simhi, Aitor Soroa, Alham~Fikri Aji, Amit Alfassy, Anna
  Rogers, Ariel~Kreisberg Nitzav, Canwen Xu, Chenghao Mou, Chris Emezue,
  Christopher Klamm, Colin Leong, Daniel van Strien, David~Ifeoluwa Adelani,
  Dragomir Radev, Eduardo~González Ponferrada, Efrat Levkovizh, Ethan Kim,
  Eyal~Bar Natan, Francesco~De Toni, Gérard Dupont, Germán Kruszewski, Giada
  Pistilli, Hady Elsahar, Hamza Benyamina, Hieu Tran, Ian Yu, Idris Abdulmumin,
  Isaac Johnson, Itziar Gonzalez-Dios, Javier de~la Rosa, Jenny Chim, Jesse
  Dodge, Jian Zhu, Jonathan Chang, Jörg Frohberg, Joseph Tobing, Joydeep
  Bhattacharjee, Khalid Almubarak, Kimbo Chen, Kyle Lo, Leandro~Von Werra, Leon
  Weber, Long Phan, Loubna~Ben allal, Ludovic Tanguy, Manan Dey, Manuel~Romero
  Muñoz, Maraim Masoud, María Grandury, Mario Šaško, Max Huang, Maximin
  Coavoux, Mayank Singh, Mike Tian-Jian Jiang, Minh~Chien Vu, Mohammad~A.
  Jauhar, Mustafa Ghaleb, Nishant Subramani, Nora Kassner, Nurulaqilla Khamis,
  Olivier Nguyen, Omar Espejel, Ona de~Gibert, Paulo Villegas, Peter Henderson,
  Pierre Colombo, Priscilla Amuok, Quentin Lhoest, Rheza Harliman, Rishi
  Bommasani, Roberto~Luis López, Rui Ribeiro, Salomey Osei, Sampo Pyysalo,
  Sebastian Nagel, Shamik Bose, Shamsuddeen~Hassan Muhammad, Shanya Sharma,
  Shayne Longpre, Somaieh Nikpoor, Stanislav Silberberg, Suhas Pai, Sydney
  Zink, Tiago~Timponi Torrent, Timo Schick, Tristan Thrush, Valentin Danchev,
  Vassilina Nikoulina, Veronika Laippala, Violette Lepercq, Vrinda Prabhu, Zaid
  Alyafeai, Zeerak Talat, Arun Raja, Benjamin Heinzerling, Chenglei Si,
  Davut~Emre Taşar, Elizabeth Salesky, Sabrina~J. Mielke, Wilson~Y. Lee,
  Abheesht Sharma, Andrea Santilli, Antoine Chaffin, Arnaud Stiegler, Debajyoti
  Datta, Eliza Szczechla, Gunjan Chhablani, Han Wang, Harshit Pandey, Hendrik
  Strobelt, Jason~Alan Fries, Jos Rozen, Leo Gao, Lintang Sutawika, M~Saiful
  Bari, Maged~S. Al-shaibani, Matteo Manica, Nihal Nayak, Ryan Teehan, Samuel
  Albanie, Sheng Shen, Srulik Ben-David, Stephen~H. Bach, Taewoon Kim, Tali
  Bers, Thibault Fevry, Trishala Neeraj, Urmish Thakker, Vikas Raunak, Xiangru
  Tang, Zheng-Xin Yong, Zhiqing Sun, Shaked Brody, Yallow Uri, Hadar Tojarieh,
  Adam Roberts, Hyung~Won Chung, Jaesung Tae, Jason Phang, Ofir Press, Conglong
  Li, Deepak Narayanan, Hatim Bourfoune, Jared Casper, Jeff Rasley, Max
  Ryabinin, Mayank Mishra, Minjia Zhang, Mohammad Shoeybi, Myriam Peyrounette,
  Nicolas Patry, Nouamane Tazi, Omar Sanseviero, Patrick von Platen, Pierre
  Cornette, Pierre~François Lavallée, Rémi Lacroix, Samyam Rajbhandari,
  Sanchit Gandhi, Shaden Smith, Stéphane Requena, Suraj Patil, Tim Dettmers,
  Ahmed Baruwa, Amanpreet Singh, Anastasia Cheveleva, Anne-Laure Ligozat, Arjun
  Subramonian, Aurélie Névéol, Charles Lovering, Dan Garrette, Deepak
  Tunuguntla, Ehud Reiter, Ekaterina Taktasheva, Ekaterina Voloshina, Eli
  Bogdanov, Genta~Indra Winata, Hailey Schoelkopf, Jan-Christoph Kalo,
  Jekaterina Novikova, Jessica~Zosa Forde, Jordan Clive, Jungo Kasai, Ken
  Kawamura, Liam Hazan, Marine Carpuat, Miruna Clinciu, Najoung Kim, Newton
  Cheng, Oleg Serikov, Omer Antverg, Oskar van~der Wal, Rui Zhang, Ruochen
  Zhang, Sebastian Gehrmann, Shachar Mirkin, Shani Pais, Tatiana Shavrina,
  Thomas Scialom, Tian Yun, Tomasz Limisiewicz, Verena Rieser, Vitaly Protasov,
  Vladislav Mikhailov, Yada Pruksachatkun, Yonatan Belinkov, Zachary Bamberger,
  Zdeněk Kasner, Alice Rueda, Amanda Pestana, Amir Feizpour, Ammar Khan, Amy
  Faranak, Ana Santos, Anthony Hevia, Antigona Unldreaj, Arash Aghagol, Arezoo
  Abdollahi, Aycha Tammour, Azadeh HajiHosseini, Bahareh Behroozi, Benjamin
  Ajibade, Bharat Saxena, Carlos~Muñoz Ferrandis, Daniel McDuff, Danish
  Contractor, David Lansky, Davis David, Douwe Kiela, Duong~A. Nguyen, Edward
  Tan, Emi Baylor, Ezinwanne Ozoani, Fatima Mirza, Frankline Ononiwu, Habib
  Rezanejad, Hessie Jones, Indrani Bhattacharya, Irene Solaiman, Irina Sedenko,
  Isar Nejadgholi, Jesse Passmore, Josh Seltzer, Julio~Bonis Sanz, Livia Dutra,
  Mairon Samagaio, Maraim Elbadri, Margot Mieskes, Marissa Gerchick, Martha
  Akinlolu, Michael McKenna, Mike Qiu, Muhammed Ghauri, Mykola Burynok, Nafis
  Abrar, Nazneen Rajani, Nour Elkott, Nour Fahmy, Olanrewaju Samuel, Ran An,
  Rasmus Kromann, Ryan Hao, Samira Alizadeh, Sarmad Shubber, Silas Wang, Sourav
  Roy, Sylvain Viguier, Thanh Le, Tobi Oyebade, Trieu Le, Yoyo Yang, Zach
  Nguyen, Abhinav~Ramesh Kashyap, Alfredo Palasciano, Alison Callahan, Anima
  Shukla, Antonio Miranda-Escalada, Ayush Singh, Benjamin Beilharz, Bo~Wang,
  Caio Brito, Chenxi Zhou, Chirag Jain, Chuxin Xu, Clémentine Fourrier,
  Daniel~León Periñán, Daniel Molano, Dian Yu, Enrique Manjavacas, Fabio
  Barth, Florian Fuhrimann, Gabriel Altay, Giyaseddin Bayrak, Gully Burns,
  Helena~U. Vrabec, Imane Bello, Ishani Dash, Jihyun Kang, John Giorgi, Jonas
  Golde, Jose~David Posada, Karthik~Rangasai Sivaraman, Lokesh Bulchandani,
  Lu~Liu, Luisa Shinzato, Madeleine~Hahn de~Bykhovetz, Maiko Takeuchi, Marc
  Pàmies, Maria~A Castillo, Marianna Nezhurina, Mario Sänger, Matthias
  Samwald, Michael Cullan, Michael Weinberg, Michiel~De Wolf, Mina Mihaljcic,
  Minna Liu, Moritz Freidank, Myungsun Kang, Natasha Seelam, Nathan Dahlberg,
  Nicholas~Michio Broad, Nikolaus Muellner, Pascale Fung, Patrick Haller, Ramya
  Chandrasekhar, Renata Eisenberg, Robert Martin, Rodrigo Canalli, Rosaline Su,
  Ruisi Su, Samuel Cahyawijaya, Samuele Garda, Shlok~S Deshmukh, Shubhanshu
  Mishra, Sid Kiblawi, Simon Ott, Sinee Sang-aroonsiri, Srishti Kumar, Stefan
  Schweter, Sushil Bharati, Tanmay Laud, Théo Gigant, Tomoya Kainuma, Wojciech
  Kusa, Yanis Labrak, Yash~Shailesh Bajaj, Yash Venkatraman, Yifan Xu, Yingxin
  Xu, Yu~Xu, Zhe Tan, Zhongli Xie, Zifan Ye, Mathilde Bras, Younes Belkada, and
  Thomas Wolf. 2023.
\newblock \href {http://arxiv.org/abs/2211.05100} {Bloom: A 176b-parameter
  open-access multilingual language model}.

\bibitem[{Yakasai(2025)}]{yakasai2025tauraruwaarshen}
S.A. Yakasai. 2025.
\newblock Tauraruwa harshen hausa jiya da yau: Kalubale da madosa.
\newblock \emph{Tauraruwa Journal of Hausa Studies}, 1(1):1--9.

\bibitem[{Yang et~al.(2024)Yang, Sun, Li, Liu, Li, Liu, Gao, and
  Huang}]{Yang20241}
Yizhe Yang, Huashan Sun, Jiawei Li, Runheng Liu, Yinghao Li, Yuhang Liu, Yang
  Gao, and Heyan Huang. 2024.
\newblock \href {https://doi.org/10.1016/j.aiopen.2024.08.001} {Mindllm:
  Lightweight large language model pre-training, evaluation and domain
  application}.
\newblock \emph{AI Open}, 5:1 – 26.

\bibitem[{Yu and Deng(2016)}]{yu2016automatic}
Dong Yu and Lin Deng. 2016.
\newblock \emph{Automatic speech recognition}, volume~1.
\newblock Springer.

\bibitem[{Yusuf et~al.(2023)Yusuf, Sarlan, Danyaro, and Rahman}]{yusuf2023fine}
Aliyu Yusuf, Aliza Sarlan, Kamaluddeen~Usman Danyaro, and Abdullahi Sani~BA
  Rahman. 2023.
\newblock Fine-tuning multilingual transformers for hausa-english sentiment
  analysis.
\newblock In \emph{2023 13th International Conference on Information Technology
  in Asia (CITA)}, pages 13--18. IEEE.

\bibitem[{Yusuf et~al.(2024)Yusuf, Sarlan, Danyaro, Rahman, and
  Abdullahi}]{yusuf2024sentiment}
Aliyu Yusuf, Aliza Sarlan, Kamaluddeen~Usman Danyaro, Abdullahi Sani~BA Rahman,
  and Mujaheed Abdullahi. 2024.
\newblock Sentiment analysis in low-resource settings: A comprehensive review
  of approaches, languages, and data sources.
\newblock \emph{IEEE Access}.

\bibitem[{Zakari et~al.(2021)Zakari, Lawal, and Abdulmumin}]{Zakari2021ASL}
Rufai~Yusuf Zakari, Zaharaddeen~Karami Lawal, and Idris Abdulmumin. 2021.
\newblock \href {https://doi.org/10.24203/ijcit.v10i4.86} {A systematic
  literature review of hausa natural language processing}.
\newblock \emph{International Journal of Computer and Information Technology
  (2279-0764)}, 10(4).

\bibitem[{Zandam et~al.(2023)Zandam, Muhammad, and
  Inuwa-Dutse}]{zandam2023online-2023}
Abubakar~Yakubu Zandam, Fatima~Adam Muhammad, and Isa Inuwa-Dutse. 2023.
\newblock Online threats detection in hausa language.
\newblock In \emph{4th Workshop on African Natural Language Processing}.

\bibitem[{Zhao et~al.(2024)Zhao, Chen, Lee, Qiu, Gao, Fan, and Lane}]{Zhao2024}
Wanru Zhao, Yihong Chen, Royson Lee, Xinchi Qiu, Yan Gao, Hongxiang Fan, and
  Nicholas~D. Lane. 2024.
\newblock \href
  {https://www.scopus.com/inward/record.uri?eid=2-s2.0-85200573076&partnerID=40&md5=994f545e0b90dd866db79d0d1a71067e}
  {Breaking physical and linguistic borders: Multilingual federated prompt
  tuning for low-resource languages}.

\bibitem[{Zhu et~al.(2023)Zhu, Zhu, Zhang, Xu, and Kong}]{zhu2023multimodal}
Linan Zhu, Zhechao Zhu, Chenwei Zhang, Yifei Xu, and Xiangjie Kong. 2023.
\newblock Multimodal sentiment analysis based on fusion methods: A survey.
\newblock \emph{Information Fusion}, 95:306--325.

\end{thebibliography}
\bibliographystyle{acl2023}

\section{Appendex}

\end{document}